\documentclass[10pt]{article}

\usepackage[T1]{fontenc}
\usepackage{graphicx}
\usepackage{todonotes}
\usepackage{hyperref}
\usepackage{libertine}
\usepackage[scaled]{beramono}
\usepackage{listings}
\usepackage{siunitx}
\usepackage{doi}
\usepackage[
	backend=biber,
	style=numeric,
	sortlocale=en_US,
	url=false, 
	doi=true,
	sortcites=true
]{biblatex}
\usepackage{fontawesome}
\usepackage{longtable}
\usepackage{subfig}

\newcommand{\wrt}{w.\,r.\,t.}

\newcommand{\Achslast}{Axle weight}
\newcommand{\Anhaenger}{Trailer}
\newcommand{\AnhaengerLKW}{Truck trailer}
\newcommand{\LKW}{Truck}
\newcommand{\GewichtLKW}{Truck weight}
\newcommand{\Omnibus}{Omnibus}
\newcommand{\Motorrad}{Motorcycle}
\newcommand{\Moped}{Moped}
\newcommand{\Fahrrad}{Cycle}
\newcommand{\FahrradMoped}{Cycle \&\ Moped}
\newcommand{\KFZ}{Power-driven}
\newcommand{\Einspurige}{Single-tracked}
\newcommand{\Reitverbot}{Riding}
\newcommand{\Fuhrwerk}{Animal-drawn}
\newcommand{\Ueberholverbot}{Overtaking}
\newcommand{\UeberholverbotLKW}{Truck overtaking}
\newcommand{\Gefahrengut}{Dangerous goods}
\newcommand{\Fussgaenger}{Pedestrian}
\newcommand{\Baustelle}{Road works}
\newcommand{\Kinder}{Children}
\newcommand{\Fussgaengeruebergang}{Pedestrian crossing}
\newcommand{\Fahrradueberfahrt}{Cyclist crossing}
\newcommand{\Schleudergefahr}{Slippery}
\newcommand{\Falschfahrer}{Wrong way driver}

\title{Comparing the Machine Readability of Traffic Sign Pictograms in Austria and Germany\thanks{This research was funded by FFG (Austrian Research Promotion Agency) under grant 879320 (SafeSign) and supported by the strategic economic research programme ``Innovatives O\"O 2020'' of the province of Upper Austria.}}
\author{Alexander Maletzky$^1$\qquad Stefan Thumfart$^1$\\[1ex] Christoph Wru\ss $^2$}
\date{\small
	$^1$ Research Unit Medical Informatics, RISC Software GmbH\\
	\textsf{$\langle$firstname.lastname$\rangle$@risc-software.at}\\[1em]
	
	$^2$ ASFINAG Service GmbH\\
	\textsf{$\langle$firstname.lastname$\rangle$@asfinag.at}
}

\begin{document}
\maketitle

\begin{abstract}
We compare the machine readability of pictograms found on Austrian and German traffic signs. To that end, we train classification models on synthetic data sets and evaluate their classification accuracy in a controlled setting. In particular, we focus on differences between currently deployed pictograms in the two countries, and a set of new pictograms designed to increase human readability. Besides other results, we find that machine-learning models generalize poorly to data sets with pictogram designs they have not been trained on. We conclude that manufacturers of advanced driver-assistance systems (ADAS) must take special care to properly address small visual differences between current and newly designed traffic sign pictograms, as well as between pictograms from different countries.
\end{abstract}

\section{Introduction}
\label{sec::Introduction}

In recent years, the number of semi-autonomous vehicles and advanced driver-assistance systems (ADAS) on our streets has been growing steadily. Even if there are still a lot of problems to be resolved before machines can eventually take over entirely, certain aspects of driving have been successfully automated already. One of them is \emph{traffic sign recognition}, which consists of \emph{detecting} and \emph{classifying} traffic signs (also called \emph{road signs}) in video frames produced by a forward-facing camera. The results of this recognition process can then be used to automatically control the speed of the vehicle, or to display the found traffic signs on the instrument panel to inform the driver about them. In either case, correctly recognizing the traffic signs is of paramount importance for avoiding potentially fatal accidents. Although state-of-the-art deep neural networks achieve near- or even super-human performance in many computer vision benchmark tasks, including traffic sign recognition~\cite{LiWang2019}, it is possible to find examples that humans can easily recognize but generally well-performing models fail to classify correctly~\cite{Szegedy2014, Sitawarin2018}. In general, it has been observed that image classification models and humans base their decisions upon fairly different aspects of images to be classified~\cite{Baker2018, Geirhos2019}, and that the former lack a great deal of the generalization capabilities humans excel at: Geirhos et al. showed that image classification models have big difficulties in classifying \emph{corrupted} images (e.\,g., by adding mild noise) if they have not been part of the data the models were trained on, quite in contrast to humans~\cite{Geirhos2018}. Dodge et al. and Azulay et al. arrive at a similar conclusion, focusing on image quality (noise, blurring, JPEG compression) and small geometric transformations (scaling, translation), respectively~\cite{Dodge2016} and~\cite{Azulay2019}. Szegedy et al. and Sitawarin et al. showed that image classifiers in general, and traffic sign classifiers in particular, are susceptible to \emph{adversarial attacks}, where images are modified in a way that is impercetible to humans but still makes the models misclassify them~\cite{Szegedy2014, Sitawarin2018}.

All this suggests that traffic sign recognition models might work well on data following the same distribution as the data they were trained on, but at the same time fail to generalize to perceptually similar albeit unseen samples. Such a situation can arise if a model is trained on the traffic signs of one country and then deployed in another country, or if existing traffic sign pictograms are replaced by a new, slightly different design. Comparing current Austrian and German traffic signs, for instance, a human who only knows Austrian traffic signs will immediately grasp the meaning of a German traffic sign when seeing it for the first time, and vice versa; whether the same applies to deep neural networks can be doubted considering the above remarks, and constitutes one of the questions we seek to answer. In addition, it would also be interesting to know whether the specific pictogram design has an impact on machine readability in the first place, ignoring transferability between different designs for the moment; this is another research question of our work. Finally, given the insights obtained from answering the first two questions we aim to conduct a qualitative evaluation of which details of a traffic sign image are especially important to a classification model, in order to derive design rules that could further improve machine readability and contribute to understanding perceptual differences between machines and humans in traffic sign recognition.

\begin{figure}[t]
	\includegraphics[width=\linewidth]{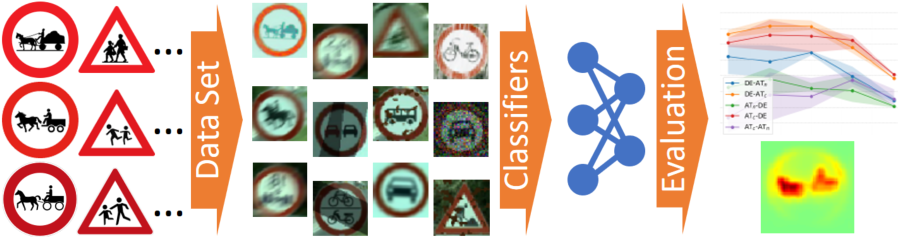}
	\caption{Overview of our experimental setup. Starting from three different sets of traffic sign pictograms we created a large collection of embedded and corrupted images (pictogram + traffic sign + background). These images are then partitioned into different groups (e.\,g., according to pictogram design or corruption intensity) and used to train classification models. Based on the models' accuracies we drew conclusions about the machine readability of the initial pictograms.}
	\label{fig::overview}
\end{figure}

For answering the questions posed above we set up a range of experiments, where we trained different \emph{traffic sign classification models} on a vast \emph{synthetic data set} with traffic signs displaying current Austrian and German pictograms, as well as proposed new Austrian pictograms.
The design of the proposed new Austrian pictograms is based on rigorous scientific methods and empirical studies, with the goal to improve human readability. Given the ever-growing presence of ADAS, however, machine readability should be taken into account as well before deploying the new design to Austrian roads.\footnote{Of course, the importance of machine- and human readability are not on par, but authorities might be reluctant to introduce a pictogram design that state-of-the-art models cannot learn to recognize. Anyway, as we will show this is not the case.} We concentrated on image classification and did not consider object detection, because changing pictograms does not affect the shape and overall appearance of a traffic sign, which is likely what a detector would mainly look at. After training the models we evaluated them on held-out test sets and compared their classification accuracy, both when applied to images featuring pictograms from the design the models were trained on and when applied to images with pictograms from a `foreign' design. We stress that the purpose of our experiments solely was to evaluate \emph{machine} readability of traffic sign pictograms; we did not intend to evaluate \emph{human} readability by any means and therefore refrain from making any claims about it throughout the paper. An overview of our experimental setup is presented in Figure~\ref{fig::overview}

Summarizing, our contributions are as follows:
\begin{itemize}
	\item We created a synthetic data set of traffic sign images with pictograms from different design. More importantly, we devised a method for systematically and automatically creating such data sets for arbitrary pictograms from any number of classes. This could prove particularly useful for training classification models on a new pictogram design before actually deploying it, i.\,e., when no real-world data is available yet.
	\item We compared the machine readability of three different pictogram designs by training and evaluating more than 100 state-of-the-art classification models, to obtain reliable results.
	\item We employed state-of-the-art techniques from explainable AI to analyze which features of an input image the models pay most attention to. Based on that we tried to identify patterns that can help to develop design rules for improving the machine readability of traffic sign pictograms.
\end{itemize}

The rest of this paper is organized as follows: Section~\ref{sec::Related} briefly reviews some related work, Section~\ref{sec::Methods} explains our methods and experimental setup in detail, and Section~\ref{sec::Results} presents our main results. Section~\ref{sec::Discussion} discusses and summarizes our results and lists potential directions for future research.

\subsection{Related Work}
\label{sec::Related}

There exists a large body of scientific work regarding the automatic detection and classification of traffic signs in real-world, as well as synthetic, data sets. One of the most widely used real-world data sets is the German Traffic Sign Recognition Benchmark (GTSRB)~\cite{Stallkamp2012} for classifying small image patches extracted from traffic scenes into one of 43 classes. Similar data sets exist for traffic signs from other countries and territories~\cite{Mogelmose2012, GamezSerna2018, Zhu2016, Tabernik2020, Timofte2014, Grigorescu2003, Belaroussi2010, Larsson2011, Yang2016}. IceVisionSet~\cite{Pavlov2019} is a data set of Russian traffic scenes, which in contrast to the other mentioned data sets was acquired entirely in winter and hence features different weather conditions. Mapillary~\cite{Mapillary} constitutes one of the largest and most diverse real-word data sets, with more than 100,000 annotated images from 6 continents. Please refer to~\cite{Temel2019} for an extensive overview of publicly available traffic sign/scene data sets. We could not make use of any real-world data set, because on the one hand no real-world data exists for the proposed new Austrian pictograms, and on the other hand a systematic, unbiased comparison of different pictogram designs can hardly be realized on real-word data.

Closer to the kind of data set we are using in our experiments are partly synthetic data sets, where photographs or video frames of full traffic scenes are augmented with ultra-realistic weather effects (rain, snow, fog)~\cite{Sakaridis2018, Bernuth2019, Volk2019}. In addition to these partly synthetic data sets there also exist data bases of fully synthetic 3D renderings of traffic scenes under varying (weather) conditions~\cite{Cognata, Ros2016, AI.Reverie, Anyverse, CVEDIA}. All these data sets have in common that they are better suited for object \emph{detection} tasks, though.

The same applies to~\cite{Temel2019, Michaelis2019}, where real-world traffic scenes are systematically modified by adding weather effects and other types of corruptions, to evaluate how well traffic sign detectors work under such `challenging conditions'. On the one hand this resembles the approach we take in our experiments, but on the other hand the main goal of the cited works is to compare different corruption types, not traffic signs or pictograms.

Recently, Berghoff et al. evaluated the robustness of classification models trained on GTSRB~\cite{Stallkamp2012} \wrt\ image corruptions~\cite{Berghoff2021}. Although the authors pursued a similarly systematic approach to creating corrupted images as we do in our work, their primary goal was to investigate how well models can handle these types of corruptions. Similarly, Hendrycks et al. created a corrupted and perturbed version of ImageNet~\cite{Deng2009}, where, as in our case, the goal is to \emph{classify} images rather than detect objects~\cite{Hendrycks2019}. The methods employed there for corrupting images are similar to ours. ImageNet, however, is a general image data base without any particular focus on traffic signs, and the goal of~\cite{Hendrycks2019} is to evaluate the performance of classification models in general, comparing models trained on `clean' images to models trained on corrupted versions thereof.

To the best of our knowledge, no systematic comparison of the machine readability (including both detection and classification) of different traffic sign (pictogram) designs exists in the literature.

\section{Methods}
\label{sec::Methods}

\subsection{Class Selection}
\label{sec::ClassSelection}

The selection of the traffic sign classes for our experiments was motivated by two main considerations: (i) Compare current Austrian pictograms with proposed new pictograms, which limits the set of classes to those where new pictogram designs are available, and (ii) focus on classes that are particularly relevant for cars, whose misclassification by a driving assistance system could have serious consequences. Concerning the availability of new pictograms, their designer\footnote{Stefan Egger, \url{https://visys.pro/}.}
kindly provided us with 37 $100\times 100$ pixel images, each corresponding to one traffic sign class. They consist of 20 prohibitory signs, six warning signs, seven mandatory signs, and four informational signs. Since the mandatory- and informational signs are mainly relevant for pedestrians and cyclists, marking footpaths, cycle tracks and crossings, we opted to exclude them from our experiments. From the 20 prohibitory signs we excluded two because their current Austrian versions are not available on the official web page~\cite{StVO-AT}, and used the remaining 18+6=\textbf{24 classes} for our experiments. See Appendix~\ref{sec::TSList} for the complete list.

\begin{figure}
	\centering
	\subfloat[Synthesized classes.]{
	\begin{tabular}{c@{ = }l}
		\raisebox{-0.4\totalheight}{\includegraphics[width=0.1\linewidth]{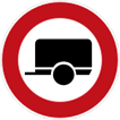}} &
		\raisebox{-0.4\totalheight}{\includegraphics[width=0.1\linewidth]{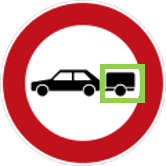}}\\
		
		\raisebox{-0.4\totalheight}{\includegraphics[width=0.1\linewidth]{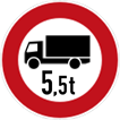}} &
		\raisebox{-0.4\totalheight}{\includegraphics[width=0.1\linewidth]{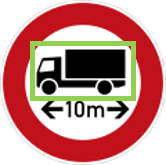}} + 
		\raisebox{-0.4\totalheight}{\includegraphics[width=0.1\linewidth]{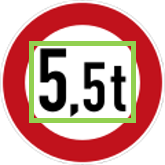}}
	\end{tabular}
	}
\qquad
	\subfloat[Replaced classes.]{
	\begin{tabular}{c@{ = }p{0.35\linewidth}}
		\raisebox{-0.4\totalheight}{\includegraphics[width=0.1\linewidth]{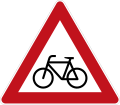}} &
		`Warning: cyclist' instead of `\Fahrradueberfahrt'\\
		
		\raisebox{-0.4\totalheight}{\includegraphics[width=0.1\linewidth]{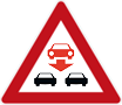}} &
		Copy of current Austrian design
	\end{tabular}
	}
	\caption{Missing German classes, and how we synthesized/replaced them.}
	\label{fig::DE_synthetic}
\end{figure}

Besides new Austrian pictograms, we also compare German traffic sign pictograms to the current Austrian design. To that end, we took the German versions of the 24 selected traffic sign classes from the official web page~\cite{VZKat}. Although the sets of Austrian and German traffic sign classes are mostly identical, there are some small differences; in particular, it turned out that four of the 24 classes do not exist in Germany. Since we did not want to further restrict the classes under consideration to the intersection of current Austrian, proposed new Austrian \emph{and} German traffic sign classes, we instead opted to hand-craft the ones missing in Germany. Figure~\ref{fig::DE_synthetic} illustrates this procedure: the pictograms for classes `\Anhaenger' and `\GewichtLKW' were synthesized by combining parts of other German pictograms, the pictogram for class `\Fahrradueberfahrt' actually corresponds to class `Warning: cyclist', and the pictogram for `\Falschfahrer' is a copy of the current Austrian design.\footnote{Class `\Falschfahrer' is an Austrian speciality.}

Some further remarks on our class selection are in place. First, we are aware that not all of the 24 classes are relevant for autonomous vehicles (think of `\Fuhrwerk', for instance). However, all of the selected classes either belong to category `Prohibitory' or `Warning', which \emph{are} particularly relevant for cars in general, and a classification model that frequently mistakes one of the not-so-relevant-classes with a highly relevant class might still be problematic. Therefore, we decided to include all available classes of the two aforementioned categories in our experiments.

Conversely, it would have also been possible to consider \emph{more} classes, even if no new pictogram design has been proposed (e.\,g., by simply copying the current design). On the one hand, this would have made the classification tasks more challenging and perhaps more `realistic', but on the other hand it would have made the systematic comparison of the different pictogram designs much harder. A large number of identical pictograms in two sets that are compared adds bias to the comparison that must be addressed properly. We leave this as future work.

\subsection{Creating the Synthetic Data Sets}
\label{sec::DatasetCreation}

We train classification models on synthetic data sets consisting of traffic sign images. For creating the images we started from high-resolution photographs of traffic scenes on Austrian highways in the year 2014 and extracted 14 patches with traffic signs. Seven of these 14 patches contain a prohibitory sign (round with red border), the other seven contain a warning sign (triangular with red border).\footnote{The source images and patches can be provided upon request.}
We then analyzed each of these 14 images \wrt\ color spectrum and perspective, obtaining parameters that allow to automatically replace the displayed pictogram by any given new pictogram in a way that makes the resulting image still look realistic. We then doubled the number of images by simply flipping them horizontally. Next, we replaced the pictograms in the 14 prohibitory signs (seven original, seven flipped) by the 18 prohibitory pictograms among the 24 traffic sign classes described in Section~\ref{sec::ClassSelection}, and did the same with the six warning signs. This procedure gives rise to a set of 336 images per pictogram design group (current Austrian, proposed new Austrian, current German), with 14 images per class. We resized these images to a uniform size of $64\times64$ pixels.

\begin{figure}
	\centering
	\begin{tabular}{ccccc}
		Contrast & Brighten & Darken & Gaussian noise & Spatter \\
		\includegraphics[width=0.15\linewidth]{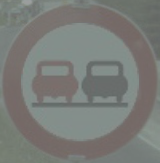} &
		\includegraphics[width=0.15\linewidth]{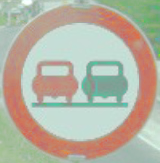} &
		\includegraphics[width=0.15\linewidth]{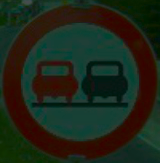} &
		\includegraphics[width=0.15\linewidth]{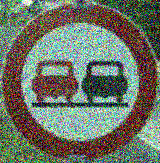} &
		\includegraphics[width=0.15\linewidth]{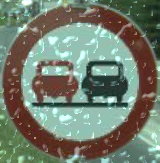} \\
		Shadow & Rain & Motion blur & Gaussian blur & Zoom blur \\
		\includegraphics[width=0.15\linewidth]{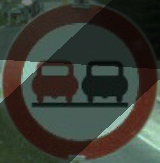} &
		\includegraphics[width=0.15\linewidth]{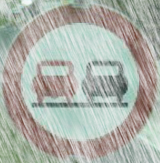} &
		\includegraphics[width=0.15\linewidth]{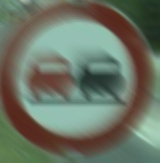} &
		\includegraphics[width=0.15\linewidth]{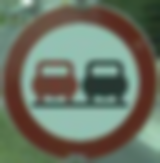} &
		\includegraphics[width=0.15\linewidth]{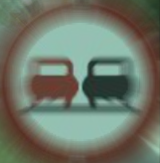}
	\end{tabular}
	\caption{Corruptions used, besides down-sampling. \emph{One} randomly chosen additional corruption is applied to every image.}
	\label{tab::augmentations}
\end{figure}

Finally, we augmented the set of 336 images by applying an arsenal of augmentation methods with varying intensities.\footnote{We used the excellent imgaug package~\cite{Jung2020} for Python for this purpose.} In particular, first one out of ten pre-selected corruption methods, like Gaussian noise, blurring, rain patterns, etc. is applied; Figure~\ref{tab::augmentations} lists all these methods. Then, the resulting images are down-sampled by first down- and then up-scaling them, to decrease their spatial resolution but keep the size of $64\times64$ pixels. The purpose of down-sampling is to simulate distance, as one of the key properties of well-designed traffic sign pictograms is their being readable from large distances. We generated 250 new images for each of the 336 clean images, with five different levels of corruption intensity (50 per level). These intensities only affect the down-sampling factor, i.\,e., a higher intensity level gives rise to more `pixelated' images. More precisely, every intensity level corresponds to a fixed interval from which the concrete down-sampling factor is sampled uniformly for every image.


\begin{figure}[t]
	\includegraphics[width=\linewidth]{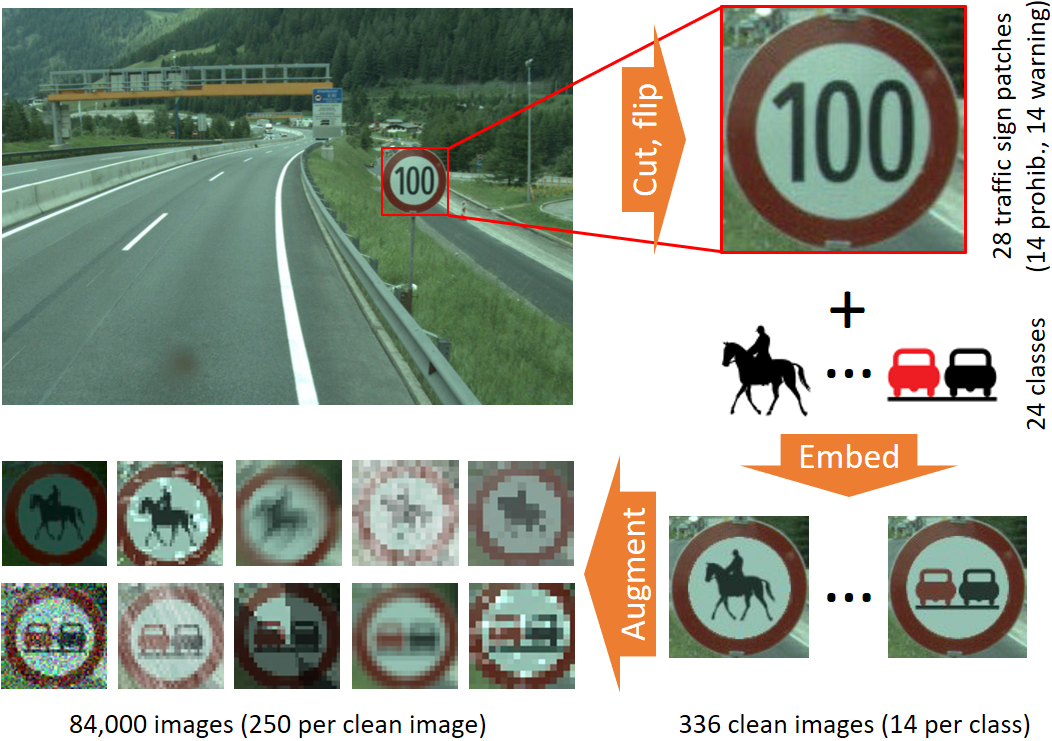}
	\caption{Data generation process for the synthetic data sets used in our experiments. This process is repeated three times for current Austrian pictograms, proposed new Austrian pictograms, and current German pictograms, yielding nine data sets with a combined total of 756,000 images.}
	\label{fig::data_generation}
\end{figure}

Figure~\ref{fig::data_generation} summarizes the whole data generation process. In the lower-left corner, two images per corruption intensity are shown, with intensity increasing from left to right. Eventually, every data set consists of 84,000 images, which are equally distributed across source patches (12,000 per patch), pictogram classes (3,500 per class) and corruption intensity (16,800 per intensity level). This, however, only corresponds to \emph{one} data set, for one group of pictograms. In our experiments we consider three distinct groups: pictograms currently deployed in Austria, proposed new Austrian pictograms, and current German pictograms. Therefore, repeating the process outlined above for each of the three groups yields three data sets with 252,000 images. In order to obtain reliable results and reduce the impact of the randomness inherent to data augmentation on our results, we repeated the entire data generation process, as well as the subsequent model training and evaluation, three times and then averaged all results over these three independent `runs'. So, in total we generated \textbf{756,000 images} for our experiments.

\begin{figure}
	\centering
	\begin{tabular}{ccc}
		\includegraphics[width=0.2\linewidth]{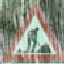} &
		\includegraphics[width=0.2\linewidth]{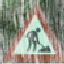} &
		\includegraphics[width=0.2\linewidth]{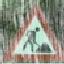}\\
		(a) current Austrian & (b) new Austrian & (c) current German
	\end{tabular}
	\caption{Comparison of corrupted images. As can be seen, the images only differ in the pictograms; even the rain pattern is exactly the same.}
	\label{tab::augmentations_old_new_DE}
\end{figure}

There is one subtlety, though: when generating the data sets for the three different pictogram groups (in the same run), we ensured that the corruptions applied to a clean image are \emph{exactly the same} for each of the three groups. This means that the three data sets only differ in their pictograms, but are otherwise identical. This is to guarantee comparability of the classification results presented in Section~\ref{sec::Results} and rule out artificial drops or increases of the classification accuracy of a model. Figure~\ref{tab::augmentations_old_new_DE} shows examples of corresponding images in the three data sets.

Finally, please note that when creating the synthetic data sets our main goal was not to produce photo-realistic images that resemble those from real-world data sets, as, for instance, in~\cite{Sakaridis2018, Bernuth2019, Volk2019}. Instead, we sought to obtain data sets that push the classification models to their limit, thereby carving out subtle differences between the machine-readability of the different pictogram groups under challenging conditions. Still, if required in a future experiment, our existing pipeline can be easily adapted to create more realistic data.

\subsection{Model Training}
\label{sec::ModelTraining}

For our experiments we trained classification models on the synthetic data sets described above, to correctly predict the class of the traffic sign shown on the input images. We did not consider object detection as a training target, because we hypothesize that traffic sign pictograms have only little influence on detection models -- after all, the shape and overall appearance of traffic signs do not change under varying pictograms, and this is likely what detectors mainly base their decisions on. Conducting experiments with detection models is therefore left as possible future work.

Models were trained separately on each of the three pictogram design groups, and also first for each of the five corruption intensities, and then jointly for all intensities. Even though this might not be the most `realistic' approach, in the sense that models actually deployed in driving assistant systems are usually trained on large data sets encompassing a big variety of traffic sign images from different countries, it helps carve out differences between the three pictogram designs more clearly.

We considered two deep neural network architectures: a small ResNet architecture with 20 layers and an input size of $64\times 64$ pixels~\cite{He2016}, and the architecture by Li and Wang with an input size of $48\times 48$ pixels~\cite{LiWang2019}. The latter was a natural choice for our experiments, since it is particularly successful in traffic sign classification.\footnote{It represents the state-of-the-art on the GTSRB data set, with $99.66\%$ test accuracy.} Putting everything together, we end up with 36 classification models per run: two models (ResNet, Li-Wang) per pictogram design group (three in total) and per corruption intensity (six in total; five separate intensity levels, plus the overall data set comprising \emph{all} images regardless of the corruption intensity). Multiplying this with the number of independent runs yields a combined total of \textbf{108 models}.

\begin{figure}[t]
	\includegraphics[width=\linewidth]{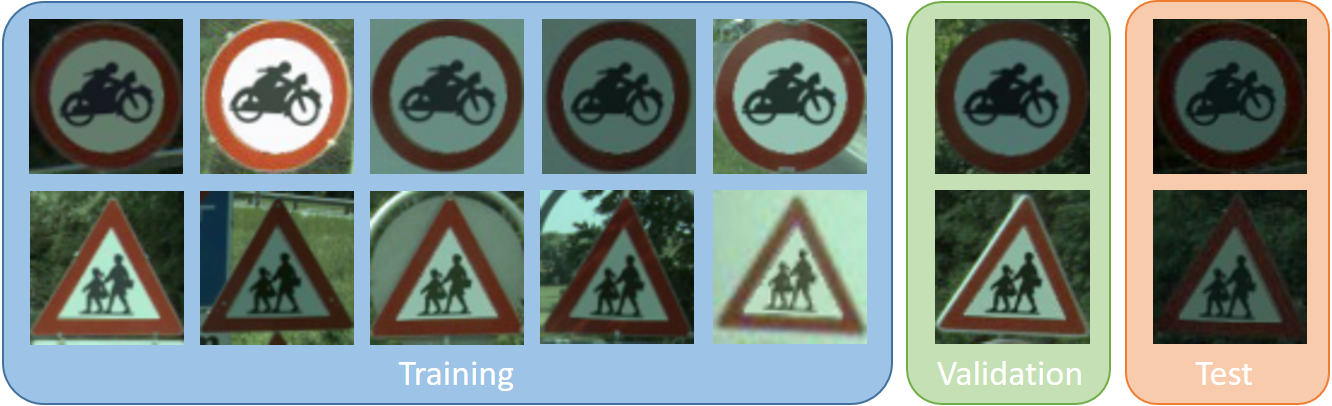}
	\caption{Splits into training-, validation- and test sets, on the image patch level. This split is identical for all traffic sign classes, pictogram designs, corruption intensities and runs. Note that the round gray shape visible in the bottom-third training image is the background of a different sign behind the warning sign, and that the flipped version of every source image is contained in the same split as the image it originates from (not depicted here).}
	\label{fig::split}
\end{figure}

All models were trained following to the same strategy. First, the available data set was split into training-, validation- and test set, in a way that puts corresponding images of different pictogram design groups in the same set, to allow for a fair comparison of the resulting models. In particular, since every corrupted image ultimately stems from one out of 14 source image patches (disregarding flipped patches), we fixed two of these patches as the `validation patches' and two as the `test patches' (one per sign shape in each case). The remaining ten patches form the `training patches'. Therefore, the training-, validation- and test images for each run and corruption intensity only differ in the pictograms between the three pictogram groups, but are otherwise identical (cf. Figure~\ref{tab::augmentations_old_new_DE}). The flipped version of every source image patch is contained in the same split as the image it originates from. The splitting strategy is summarized in Figure~\ref{fig::split}. Please note that even though corruptions are identical for the pictogram designs, they are not necessarily identical for the various image patches. Thus, it is very likely that the test sets contain images that were corrupted in a different way than all training- and validation images. We did not, however, explicitly exclude a whole corruption type from training to assess how well the models can adapt to unseen corruptions during testing, as in~\cite{Michaelis2019}. This is because we are interested in comparing \emph{pictogram designs}, not corruption types or models.

We trained the models for 60 epochs, using the Adam optimizer~\cite{Kingma2014} with an initial learning rate of $0.001$, $\beta_1=0.9$ and $\beta_2=0.999$ (these are the defaults in Keras). The learning rate is reduced by $80\%$ whenever the validation loss does not improve for ten epochs. In the end, the trained weights of the epoch with the smallest validation loss are taken. Both training- and validation accuracy plateau after only a few ($<10$) epochs in each case, so training for a total of 60 epochs is certainly sufficient.

\subsection{Evaluation Strategy}
\label{sec::EvaluationStrategy}

After training we evaluated all models on the held-out test sets, using the overall classification accuracy as the main metric of interest. This choice is motivated by the fact that all our data sets are perfectly balanced \wrt\ the 24 classes \emph{by construction}, making accuracy a feasible performance metric for comparing different models. As is common practice in this scientific area, confusion between different classes is treated uniformly. Putting less weight on confusion between semantically similar classes (e.\,g., `\Fussgaengeruebergang' and `\Fahrradueberfahrt') could be an interesting direction for future research, though.

First, every model is evaluated on its `own' test set, i.\,e., the held-out part of the data set it was trained on. Due to the uniform construction of training-, validation- and test sets, the performance scores thus obtained are feasible for comparing the quality of different models. This, in particular, also applies to models trained on different pictogram design groups, which is the main objective of our experiments: comparing the machine-readability of different pictogram designs. In the remainder, we will label evaluations of the `current Austrian' models\footnote{That is, models trained on current Austrian pictograms.} on their own test sets as `AT$_c$-AT$_c$', and likewise label evaluations of the `proposed new Austrian' and the `current German' models as `AT$_n$-AT$_n$' and `DE-DE', respectively.

In addition, however, models are evaluated on other, `foreign' test sets as well:
\begin{itemize}
	\item Models trained on the current Austrian pictograms are evaluated on proposed new Austrian pictograms and German pictograms. The first is to investigate how well existing classification models can be expected to generalize to new pictograms, and the second to investigate how well they generalize to German pictograms.
	
	In the remainder, we will label these evaluations as `AT$_c$-AT$_n$' and `AT$_c$-DE', respectively.
	\item Models trained on the proposed new Austrian pictograms are evaluated on German pictograms (labeled as `AT$_n$-DE' in the remainder).
	\item Models trained on the current German pictograms are evaluated on current and proposed new Austrian pictograms (labeled as `DE-AT$_c$' and `DE-AT$_n$', respectively).
\end{itemize}
In each case, every model is evaluated on a foreign test set of the same corruption intensity and run as its training set. For instance, a model trained on current Austrian pictograms corrupted at intensity level 5 is never evaluated on proposed new Austrian pictograms corrupted at intensity level 4. Moreover, as a general rule, models are \emph{never} compared across different runs; instead, all results are averaged over the three runs.

Please note that models trained on new Austrian pictograms are not evaluated on current Austrian pictograms, as this does not represent a realistic scenario. We also conducted preliminary experiments on the effect of training models on multiple pictogram designs simultaneously; details can be found in Appendix~\ref{sec::Mixed}.

\section{Results}
\label{sec::Results}

We focus on the results of the Li-Wang models~\cite{LiWang2019} in this section and refer the interested reader to Appendix~\ref{sec::ResNet} for a thorough discussion of the results obtained from the ResNet models. Anyway, the results of the ResNet models exhibit the same overall tendency as the Li-Wang models, meaning that the discussion in Section~\ref{sec::Discussion} applies equally to both model architectures.

Recall from Section~\ref{sec::Methods} that all results represent the average over three runs (model training and -evaluation) on independent data sets.

\subsection{Overall Results}
\label{sec::OverallResults}

\begin{table}[t]
	\centering
	\begin{tabular}{l@{\qquad}rrrrrr}
		& 1 & 2 & 3 & 4 & 5 & All\\\hline\hline
		AT$_c$-AT$_c$ & 99.58 & 99.40 & 98.04 & \textbf{97.61} & 94.75 & \textbf{98.89}\\
		AT$_n$-AT$_n$ & 99.60 & \textbf{99.53} & \textbf{98.92} & 96.96 & 94.56 & 98.68\\
		DE-DE & \textbf{99.72} & 99.38 & 98.78 & 97.13 & \textbf{94.86} & 98.85\\\hline
		AT$_c$-AT$_n$ & 68.21 & 71.93 & 71.65 & 74.26 & 71.26 & 80.18\\
		AT$_c$-DE & 80.24 & 81.46 & 81.31 & 80.63 & 75.11 & 83.94\\
		AT$_n$-DE & 73.10 & 74.42 & 72.93 & 72.60 & 70.11 & 75.33\\
		DE-AT$_c$ & 81.63 & 82.90 & 82.83 & 79.46 & 74.51 & 82.03\\
		DE-AT$_n$ & 78.04 & 77.18 & 78.69 & 74.86 & 71.04 & 77.35\\
	\end{tabular}
	\caption{Classification accuracy (\%) of the Li-Wang models trained in our experiments. Columns correspond to corruption intensity levels. Shown is the average accuracy over three runs, with the top accuracy per corruption intensity in \textbf{bold}.}
	\label{tab::acc}
\end{table}

\begin{figure}
	\centering
	\footnotesize
	\begin{tabular}{cc}
		\includegraphics[width=0.45\linewidth]{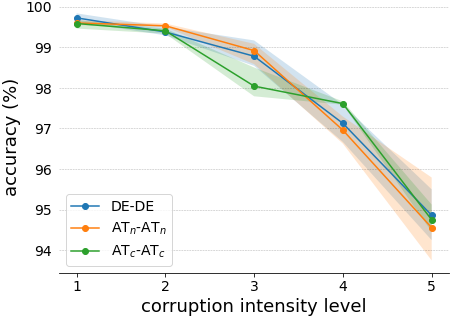} & \includegraphics[width=0.45\linewidth]{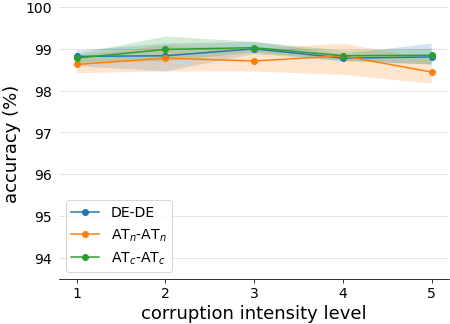}\\
		(a) `own' test sets, separate models per intensity & (b) `own' test sets, one model for all intensities\\[1em]
		\includegraphics[width=0.45\linewidth]{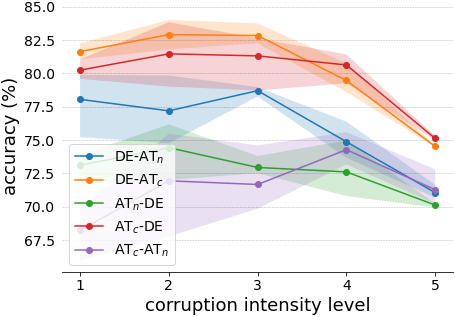} & \includegraphics[width=0.45\linewidth]{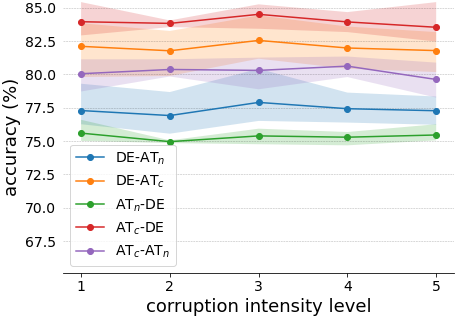}\\
		(c) `foreign' test sets, separate models per intensity & (d) `foreign' test sets, one model for all intensities
	\end{tabular}
	\caption{Classification accuracy plotted against corruption intensity. The semi-transparent areas indicate the minimum/maximum over the three runs. In (b) and (d), even though the models were trained on the full data set with all corruption intensities, the classification accuracy is again shown for each corruption intensity.}
	\label{fig::acc}
\end{figure}

Table~\ref{tab::acc} contains the final classification accuracies of all experiments and corruption intensities, and Figure~\ref{fig::acc} visualizes them as line plots where the classification accuracy is plotted against the corruption intensity level. In Figure~\ref{fig::acc}~(a) one can clearly observe that the quality of the classification models decreases as the corruption intensity increases, as expected. Interestingly, however, this effect is diminished or even reversed when one single model is trained on the whole data set, with images from all corruption intensities combined (Figure~\ref{fig::acc}~(b) and~(d)). We hypothesize that the models profit from less aggressively down-sampled images in the training set for classifying aggressively down-sampled images in the test set. It seems that by training on a more diverse data set the resulting models become more robust against severe corruptions, but at the same time fail to reach top-performance on mildly corrupted images. A similar effect has been observed for adversarially trained image classification models~\cite{Tsipras2019}. Overall, one can see that there is hardly any difference in the classification accuracy of the models between the three pictogram design groups (Figure~\ref{fig::acc}~(a) and~(b)).

In addition, one can also see very clearly that the classification accuracy of every model drops significantly when evaluated on a `foreign' test set, with different (albeit similar) pictograms. In fact, the difference between current and proposed new Austrian pictograms seems to be more pronounced than the difference between current Austrian and German pictograms (Figure~\ref{fig::acc}~(c) and~(d)). Models trained on German pictograms generalize only poorly to new Austrian pictograms, and vice versa; this is particularly interesting, since intuitively the design of the new Austrian pictograms resembles the German design much closer than the current Austrian design does, especially \wrt\ stroke width and level of detail.\footnote{See Appendix~\ref{sec::TSList} for a complete list.}

\subsection{Per-Class Results for Own Test Sets}
\label{sec::ClassResultsOwn}

\begin{table}[t]
	\renewcommand{\arraystretch}{2}
	\subfloat[AT$_c$-AT$_c$]{
	\begin{tabular}{c@{ $\rightarrow$ }c@{\quad}r}
		\raisebox{-0.35\totalheight}{\includegraphics[width=0.07\linewidth]{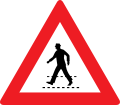}} & \raisebox{-0.35\totalheight}{\includegraphics[width=0.07\linewidth]{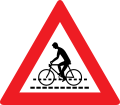}} & 1.87\%\\
		
		\raisebox{-0.35\totalheight}{\includegraphics[width=0.07\linewidth]{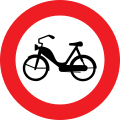}} & \raisebox{-0.35\totalheight}{\includegraphics[width=0.07\linewidth]{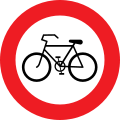}} & 1.07\%\\
		
		\raisebox{-0.35\totalheight}{\includegraphics[width=0.07\linewidth]{img/sign_classes/old/Fahrradueberfahrt.png}} & \raisebox{-0.35\totalheight}{\includegraphics[width=0.07\linewidth]{img/sign_classes/old/Fussgaengeruebergang.png}} & 0.93\%\\
		
		\raisebox{-0.35\totalheight}{\includegraphics[width=0.07\linewidth]{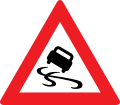}} & \raisebox{-0.35\totalheight}{\includegraphics[width=0.07\linewidth]{img/sign_classes/old/Fussgaengeruebergang.png}} & 0.73\%\\
		
		\raisebox{-0.35\totalheight}{\includegraphics[width=0.07\linewidth]{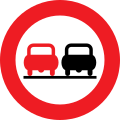}} & \raisebox{-0.35\totalheight}{\includegraphics[width=0.07\linewidth]{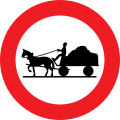}} & 0.60\%
	\end{tabular}
	}
	\hfill
	\subfloat[AT$_n$-AT$_n$]{
	\begin{tabular}{c@{ $\rightarrow$ }c@{\quad}r}
		\raisebox{-0.35\totalheight}{\includegraphics[width=0.07\linewidth]{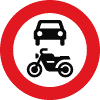}} & \raisebox{-0.35\totalheight}{\includegraphics[width=0.07\linewidth]{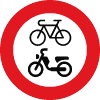}} & 1.60\%\\
		
		\raisebox{-0.35\totalheight}{\includegraphics[width=0.07\linewidth]{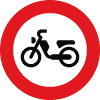}} & \raisebox{-0.35\totalheight}{\includegraphics[width=0.07\linewidth]{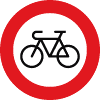}} & 1.07\%\\
		
		\raisebox{-0.35\totalheight}{\includegraphics[width=0.07\linewidth]{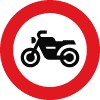}} & \raisebox{-0.35\totalheight}{\includegraphics[width=0.07\linewidth]{img/sign_classes/new/Moped.png}} & 1.07\%\\
		
		\raisebox{-0.35\totalheight}{\includegraphics[width=0.07\linewidth]{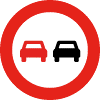}} & \raisebox{-0.35\totalheight}{\includegraphics[width=0.07\linewidth]{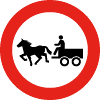}} & 0.67\%\\
		
		\raisebox{-0.35\totalheight}{\includegraphics[width=0.07\linewidth]{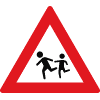}} & \raisebox{-0.35\totalheight}{\includegraphics[width=0.07\linewidth]{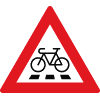}} & 0.53\%
	\end{tabular}
	}
	\hfill
	\subfloat[DE-DE]{
	\begin{tabular}{c@{ $\rightarrow$ }c@{\quad}r}
		\raisebox{-0.35\totalheight}{\includegraphics[width=0.07\linewidth]{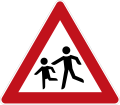}} & \raisebox{-0.35\totalheight}{\includegraphics[width=0.07\linewidth]{img/sign_classes/DE/Fahrradueberfahrt.png}} & 1.47\%\\
		
		\raisebox{-0.35\totalheight}{\includegraphics[width=0.07\linewidth]{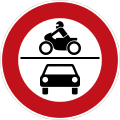}} & \raisebox{-0.35\totalheight}{\includegraphics[width=0.07\linewidth]{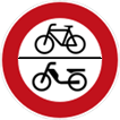}} & 0.93\%\\
		
		\raisebox{-0.35\totalheight}{\includegraphics[width=0.07\linewidth]{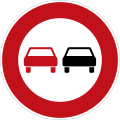}} & \raisebox{-0.35\totalheight}{\includegraphics[width=0.07\linewidth]{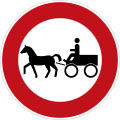}} & 0.93\%\\
		
		\raisebox{-0.35\totalheight}{\includegraphics[width=0.07\linewidth]{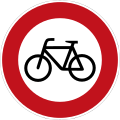}} & \raisebox{-0.35\totalheight}{\includegraphics[width=0.07\linewidth]{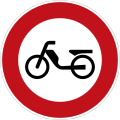}} & 0.67\%\\
		
		\raisebox{-0.35\totalheight}{\includegraphics[width=0.07\linewidth]{img/sign_classes/DE/Moped.png}} & \raisebox{-0.35\totalheight}{\includegraphics[width=0.07\linewidth]{img/sign_classes/DE/Fahrrad.png}} & 0.67\%
	\end{tabular}
	}
	\caption{Frequent confusion of the models trained on the complete data sets. The numbers on the right are the percentages of samples belonging to the class on the left-hand-side of the arrows, which are misclassified as the class on the right-hand-side of the arrows.}
	\label{tab::confusion_own}
\end{table}

\begin{table}
	\centering
	\begin{tabular}{crrrrrr}
		& \multicolumn{2}{c}{AT$_c$-AT$_c$} & \multicolumn{2}{c}{AT$_n$-AT$_n$} & \multicolumn{2}{c}{DE-DE}\\
		& $\rightarrow$ & $\leftarrow$ & $\rightarrow$ & $\leftarrow$ & $\rightarrow$ & $\leftarrow$\\\hline
		
		\raisebox{-0.35\totalheight}{\includegraphics[width=0.07\linewidth]{img/sign_classes/old/Moped.png}} vs. \raisebox{-0.35\totalheight}{\includegraphics[width=0.07\linewidth]{img/sign_classes/old/Fahrrad.png}}
		& 1.07 & 0.27 & 1.07 & 0.47 & 0.67 & 0.67\\
		
		\raisebox{-0.35\totalheight}{\includegraphics[width=0.07\linewidth]{img/sign_classes/old/Ueberholverbot.png}} vs. \raisebox{-0.35\totalheight}{\includegraphics[width=0.07\linewidth]{img/sign_classes/old/Fuhrwerk.png}}
		& 0.60 & 0.13 & 0.67 & 0.20 & 0.93 & 0.13\\
		
		\raisebox{-0.35\totalheight}{\includegraphics[width=0.07\linewidth]{img/sign_classes/old/Fussgaengeruebergang.png}} vs. \raisebox{-0.35\totalheight}{\includegraphics[width=0.07\linewidth]{img/sign_classes/old/Fahrradueberfahrt.png}}
		& 1.87 & 0.93 & 0.27 & 0.00 & 0.13 & 0.47\\
		
		\raisebox{-0.35\totalheight}{\includegraphics[width=0.07\linewidth]{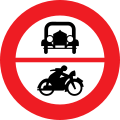}} vs. \raisebox{-0.35\totalheight}{\includegraphics[width=0.07\linewidth]{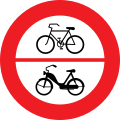}}
		& 0.40 & 0.13 & 1.6 & 0.40 & 0.93 & 0.27\\
		
		\raisebox{-0.35\totalheight}{\includegraphics[width=0.07\linewidth]{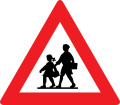}} vs. \raisebox{-0.35\totalheight}{\includegraphics[width=0.07\linewidth]{img/sign_classes/old/Fahrradueberfahrt.png}}
		& 0.60 & 0.20 & 0.53 & 0.40 & 1.47 & 0.20
	\end{tabular}
	\caption{Confusion percentage for selected pairs of classes, in both directions. Refers to models that were trained on the complete data sets. Note that even though only Austrian pictograms are shown in the table, all models are evaluated on the pictogram design they were trained on.}
	\label{tab::confusion_own_detailed}
\end{table}

Table~\ref{tab::confusion_own} lists the pairs of traffic sign classes the models confuse most often, and Table~\ref{tab::confusion_own_detailed} highlights five pairs of classes that at least one of the models confuses frequently. As one can see, all models quite often confuse `\Moped' and `\Fahrrad', misclassify `\Kinder' as `\Fahrradueberfahrt', and misclassify `\Ueberholverbot' as `\Fuhrwerk' despite the fact that the latter class lacks the red color that is present in the former class. They occasionally misclassify `\KFZ` as `\FahrradMoped', although the models trained on proposed new Austrian and German pictograms tend to have greater difficulty than the models trained on current Austrian pictograms. Furthermore, the models trained on the proposed new Austrian pictograms frequently misclassify `\Motorrad' as `\Moped', in contrast to the other models. This can possibly be explained by the fact that both the current Austrian and German `\Motorrad' pictograms look quite different than their `\Moped' counterparts: the current Austrian version heads into the other direction, whereas the German design features a person riding the motorcycle. On the other hand, the models trained on current Austrian pictograms quite often confuse `\Fussgaengeruebergang' and `\Fahrradueberfahrt', which the other models do not; in particular, the models trained on new Austrian pictograms \emph{never} misclassify `\Fahrradueberfahrt' as `\Fussgaengeruebergang' (the $0.00$ in the corresponding row of Table~\ref{tab::confusion_own_detailed} has not been rounded) and rarely make a mistake in the other direction. Complete per-class accuracy tables and full confusion matrices can be found in Appendix~\ref{sec::DetailedClass}.

%

\subsection{Per-Class Results for Foreign Test Sets}
\label{sec::ClassResultsForeign}

\begin{table}[t]
	\centering
	\renewcommand{\arraystretch}{2}
	\subfloat[AT$_c$-AT$_n$]{
		\begin{tabular}{c@{/}c@{ $\rightarrow$ }c@{\quad}r}
			\raisebox{-0.35\totalheight}{\includegraphics[width=0.07\linewidth]{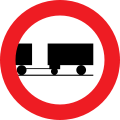}} &
			\raisebox{-0.35\totalheight}{\includegraphics[width=0.07\linewidth]{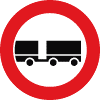}} & \raisebox{-0.35\totalheight}{\includegraphics[width=0.07\linewidth]{img/sign_classes/old/Fuhrwerk.png}} & 68.7\%\\
			
			\raisebox{-0.35\totalheight}{\includegraphics[width=0.07\linewidth]{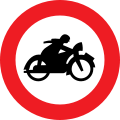}} &
			\raisebox{-0.35\totalheight}{\includegraphics[width=0.07\linewidth]{img/sign_classes/new/Motorrad.png}} & \raisebox{-0.35\totalheight}{\includegraphics[width=0.07\linewidth]{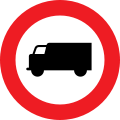}} & 54.1\%\\
			
			\raisebox{-0.35\totalheight}{\includegraphics[width=0.07\linewidth]{img/sign_classes/old/Kinder.png}} &
			\raisebox{-0.35\totalheight}{\includegraphics[width=0.07\linewidth]{img/sign_classes/new/Kinder.png}} & \raisebox{-0.35\totalheight}{\includegraphics[width=0.07\linewidth]{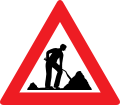}} & 29.3\%\\
			
			\raisebox{-0.35\totalheight}{\includegraphics[width=0.07\linewidth]{img/sign_classes/old/Fahrradueberfahrt.png}} &
			\raisebox{-0.35\totalheight}{\includegraphics[width=0.07\linewidth]{img/sign_classes/new/Fahrradueberfahrt.png}} & \raisebox{-0.35\totalheight}{\includegraphics[width=0.07\linewidth]{img/sign_classes/old/Kinder.png}} & 28.4\%\\
			
			\raisebox{-0.35\totalheight}{\includegraphics[width=0.07\linewidth]{img/sign_classes/old/Fahrradueberfahrt.png}} &
			\raisebox{-0.35\totalheight}{\includegraphics[width=0.07\linewidth]{img/sign_classes/new/Fahrradueberfahrt.png}} & \raisebox{-0.35\totalheight}{\includegraphics[width=0.07\linewidth]{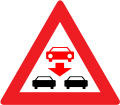}} & 27.8\%
		\end{tabular}
	}
	\hfill
	\subfloat[DE-AT$_c$]{
		\begin{tabular}{c@{/}c@{ $\rightarrow$ }c@{\quad}r}
			\raisebox{-0.35\totalheight}{\includegraphics[width=0.07\linewidth]{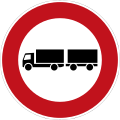}} &
			\raisebox{-0.35\totalheight}{\includegraphics[width=0.07\linewidth]{img/sign_classes/old/Anhaenger_LKW.png}} & \raisebox{-0.35\totalheight}{\includegraphics[width=0.07\linewidth]{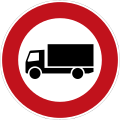}} & 87.6\%\\
			
			\raisebox{-0.35\totalheight}{\includegraphics[width=0.07\linewidth]{img/sign_classes/DE/Moped.png}} &
			\raisebox{-0.35\totalheight}{\includegraphics[width=0.07\linewidth]{img/sign_classes/old/Moped.png}} & \raisebox{-0.35\totalheight}{\includegraphics[width=0.07\linewidth]{img/sign_classes/DE/Fahrrad.png}} & 55.3\%\\
			
			\raisebox{-0.35\totalheight}{\includegraphics[width=0.07\linewidth]{img/sign_classes/DE/Fahrradueberfahrt.png}} &
			\raisebox{-0.35\totalheight}{\includegraphics[width=0.07\linewidth]{img/sign_classes/old/Fahrradueberfahrt.png}} & \raisebox{-0.35\totalheight}{\includegraphics[width=0.07\linewidth]{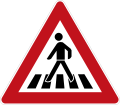}} & 51.6\%\\
			
			\raisebox{-0.35\totalheight}{\includegraphics[width=0.07\linewidth]{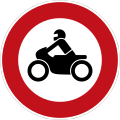}} &
			\raisebox{-0.35\totalheight}{\includegraphics[width=0.07\linewidth]{img/sign_classes/old/Motorrad.png}} & \raisebox{-0.35\totalheight}{\includegraphics[width=0.07\linewidth]{img/sign_classes/DE/LKW.png}} & 27.9\%\\
			
			\raisebox{-0.35\totalheight}{\includegraphics[width=0.07\linewidth]{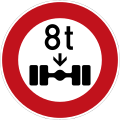}} &
			\raisebox{-0.35\totalheight}{\includegraphics[width=0.07\linewidth]{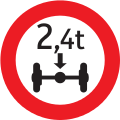}} & \raisebox{-0.35\totalheight}{\includegraphics[width=0.07\linewidth]{img/sign_classes/DE/Fahrrad_Moped.png}} & 27.1\%
		\end{tabular}
	}
	\hfill
	\subfloat[DE-AT$_n$]{
		\begin{tabular}{c@{/}c@{ $\rightarrow$ }c@{\quad}r}
			\raisebox{-0.35\totalheight}{\includegraphics[width=0.07\linewidth]{img/sign_classes/DE/Kinder.png}} &
			\raisebox{-0.35\totalheight}{\includegraphics[width=0.07\linewidth]{img/sign_classes/new/Kinder.png}} & \raisebox{-0.35\totalheight}{\includegraphics[width=0.07\linewidth]{img/sign_classes/DE/Fahrradueberfahrt.png}} & 95.3\%\\
			
			\raisebox{-0.35\totalheight}{\includegraphics[width=0.07\linewidth]{img/sign_classes/DE/Motorrad.png}} &
			\raisebox{-0.35\totalheight}{\includegraphics[width=0.07\linewidth]{img/sign_classes/new/Motorrad.png}} & \raisebox{-0.35\totalheight}{\includegraphics[width=0.07\linewidth]{img/sign_classes/DE/LKW.png}} & 66.3\%\\
			
			\raisebox{-0.35\totalheight}{\includegraphics[width=0.07\linewidth]{img/sign_classes/DE/Achslast.png}} &
			\raisebox{-0.35\totalheight}{\includegraphics[width=0.07\linewidth]{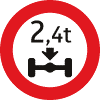}} & \raisebox{-0.35\totalheight}{\includegraphics[width=0.07\linewidth]{img/sign_classes/DE/Fahrrad_Moped.png}} & 54.7\%\\
			
			\raisebox{-0.35\totalheight}{\includegraphics[width=0.07\linewidth]{img/sign_classes/DE/Anhaenger_LKW.png}} &
			\raisebox{-0.35\totalheight}{\includegraphics[width=0.07\linewidth]{img/sign_classes/new/Anhaenger_LKW.png}} & \raisebox{-0.35\totalheight}{\includegraphics[width=0.07\linewidth]{img/sign_classes/DE/LKW.png}} & 41.8\%\\
			
			\raisebox{-0.35\totalheight}{\includegraphics[width=0.07\linewidth]{img/sign_classes/DE/KFZ.png}} &
			\raisebox{-0.35\totalheight}{\includegraphics[width=0.07\linewidth]{img/sign_classes/new/KFZ.png}} & \raisebox{-0.35\totalheight}{\includegraphics[width=0.07\linewidth]{img/sign_classes/DE/Fahrrad_Moped.png}} & 38.7\%
		\end{tabular}
	}
	\caption{Frequent confusion of the models trained on the complete data sets, when evaluated on `foreign' test sets. The numbers on the right are the percentages of samples belonging to the class on the left-hand-side of the arrows, which are misclassified as the class on the right-hand-side of the arrows. For better comparison, both training- and evaluation pictograms are shown on the left-hand-side of the arrows.}
	\label{tab::confusion_foreign}
\end{table}

\begin{table}
	\centering
	\begin{tabular}{crrrrrr}
		& \multicolumn{2}{c}{AT$_c$-AT$_n$} & \multicolumn{2}{c}{DE-AT$_c$} & \multicolumn{2}{c}{DE-AT$_n$}\\
		& $\rightarrow$ & $\leftarrow$ & $\rightarrow$ & $\leftarrow$ & $\rightarrow$ & $\leftarrow$\\\hline
		
		\raisebox{-0.35\totalheight}{\includegraphics[width=0.07\linewidth]{img/sign_classes/old/Achslast.png}} vs. \raisebox{-0.35\totalheight}{\includegraphics[width=0.07\linewidth]{img/sign_classes/old/Fahrrad_Moped.png}}
		& 0.47 & 5.87 & 27.07 & 0.00 & 54.67 & 0.33\\
		
		\raisebox{-0.35\totalheight}{\includegraphics[width=0.07\linewidth]{img/sign_classes/old/Motorrad.png}} vs. \raisebox{-0.35\totalheight}{\includegraphics[width=0.07\linewidth]{img/sign_classes/old/LKW.png}}
		& 54.13 & 6.87 & 27.87 & 0.00 & 66.33 & 0.00\\
		
		\raisebox{-0.35\totalheight}{\includegraphics[width=0.07\linewidth]{img/sign_classes/old/Kinder.png}} vs. \raisebox{-0.35\totalheight}{\includegraphics[width=0.07\linewidth]{img/sign_classes/old/Fahrradueberfahrt.png}}
		& 7.27 & 28.40 & 10.33 & 6.53 & 95.27 & 0.53\\
		
		\raisebox{-0.35\totalheight}{\includegraphics[width=0.07\linewidth]{img/sign_classes/old/Anhaenger_LKW.png}} vs. \raisebox{-0.35\totalheight}{\includegraphics[width=0.07\linewidth]{img/sign_classes/old/LKW.png}}
		& 0.40 & 3.33 & 87.60 & 0.40 & 41.80 & 0.07\\
		
		\raisebox{-0.35\totalheight}{\includegraphics[width=0.07\linewidth]{img/sign_classes/old/Moped.png}} vs. \raisebox{-0.35\totalheight}{\includegraphics[width=0.07\linewidth]{img/sign_classes/old/Fahrrad.png}}
		& 4.47 & 3.47 & 55.33 & 2.07 & 2.73 & 1.40
	\end{tabular}
	\caption{Confusion percentage for selected pairs of classes, in both directions. Refers to models that were trained on the complete data sets. Even though only Austrian pictograms are shown in the table, all models are evaluated on the pictogram design specified in the table header.}
	\label{tab::confusion_foreign_detailed}
\end{table}

Table~\ref{tab::confusion_foreign} lists the pairs of traffic sign classes the models confuse most often if the pictogram design differs from the design they were trained on, and Table~\ref{tab::confusion_foreign_detailed} focuses on some pairs in more detail. Class `\AnhaengerLKW' seems to cause most problems: DE-AT$_c$ and DE-AT$_n$ often confuse `\AnhaengerLKW' with `\LKW'; AT$_c$-AT$_n$ hardly ever confuses these two classes, but instead misclassifies `\AnhaengerLKW' as `\Fuhrwerk', such that in the total the accuracy of `\AnhaengerLKW' drops as far as $1.13\%$, as can be seen in Table~\ref{tab::class_acc_foreign}. It can also be seen that in all evaluations `\Motorrad' is frequently misclassified as `\LKW'. This might be explainable by the three designs of `\Motorrad' differing fairly strongly (cf. Appendix~\ref{sec::TSList}), meaning that the models become uncertain about their predictions when exposed to a different design. An analogous statement applies to `\KFZ'. It is also interesting to note that even though the models most often misclassify `\Moped' as `\Fahrrad' when evaluated on their own test sets (cf. Table~\ref{tab::confusion_own}), `\Moped' is only among the five most frequently confused classes in the DE-AT$_c$ evaluation.

\begin{table}
	\centering
	\begin{tabular}{crrr}
		& AT$_c$-AT$_n$ & DE-AT$_c$ & DE-AT$_n$\\\hline
		
		\raisebox{-0.35\totalheight}{\includegraphics[width=0.07\linewidth]{img/sign_classes/old/Anhaenger_LKW.png}}
		& 11.87\% (24) & 1.13\% (24) & 15.73\% (22)\\
		
		\raisebox{-0.35\totalheight}{\includegraphics[width=0.07\linewidth]{img/sign_classes/old/Motorrad.png}}
		& 29.20\% (22) & 56.60\% (21) & 1.60\% (24)\\
		
		\raisebox{-0.35\totalheight}{\includegraphics[width=0.07\linewidth]{img/sign_classes/old/Kinder.png}}
		& 59.40\% (21) & 85.33\% (17) & 2.80\% (23)\\
		
		\raisebox{-0.35\totalheight}{\includegraphics[width=0.07\linewidth]{img/sign_classes/old/Fahrradueberfahrt.png}}
		& 28.93\% (23) & 36.27\% (23) & 89.33\% (15)\\
		
		\raisebox{-0.35\totalheight}{\includegraphics[width=0.07\linewidth]{img/sign_classes/old/KFZ.png}}
		& 64.53\% (20) & 85.80\% (15) & 48.27\% (20)\\
		
		\raisebox{-0.35\totalheight}{\includegraphics[width=0.07\linewidth]{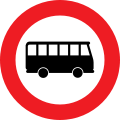}}
		& 83.87\% (15) & 76.60\% (19) & 63.73\% (19)\\
		
		\raisebox{-0.35\totalheight}{\includegraphics[width=0.07\linewidth]{img/sign_classes/old/Ueberholverbot.png}}
		& 87.47\% (14) & 82.87\% (18) & 95.60\% (13)\\
		
		\raisebox{-0.35\totalheight}{\includegraphics[width=0.07\linewidth]{img/sign_classes/old/Moped.png}}
		& 90.93\% (13) & 42.13\% (22) & 95.93\% (11)\\
		
		\raisebox{-0.35\totalheight}{\includegraphics[width=0.07\linewidth]{img/sign_classes/old/Achslast.png}}
		& 98.80\% \phantom{0}(2) & 71.33\% (20) & 38.93\% (21)\\
		
		\raisebox{-0.35\totalheight}{\includegraphics[width=0.07\linewidth]{img/sign_classes/old/Fussgaengeruebergang.png}}
		& 69.87\% (19) & 85.47\% (16) & 98.40\% \phantom{0}(2)
	\end{tabular}
	\caption{Accuracy of selected classes. Refers to models that were trained on the complete data sets. Numbers in parentheses denote the rank among all 24 classes. Even though only Austrian pictograms are shown in the table, all models are evaluated on the pictogram design indicated in the table header.}
	\label{tab::class_acc_foreign}
\end{table}

Table~\ref{tab::class_acc_foreign} lists the per-class accuracy of the models for a couple of selected classes. Although the overall performance of DE-AT$_c$ is best ($82.03\%$ accuracy; cf. Table~\ref{tab::acc}), its worst per-class accuracy of $1.13\%$, attained at class `\AnhaengerLKW', is considerably lower than the worst per-class accuracy of AT$_c$-AT$_n$ ($11.87\%$). This means that AT$_c$-AT$_n$ confuses a larger number of classes, whereas DE-AT$_c$ and DE-AT$_n$ confuse a smaller number of classes more severely.

In general, although the overall classification accuracy of all models drops considerably on foreign pictogram designs, there are blatant inter-class differences. In fact, a big deal of this drop is caused by only a few classes, namely those listed in Table~\ref{tab::class_acc_foreign}; the others are correctly classified most of the time.


\subsection{Qualitative Explanations of the Models' Predictions}
\label{sec::XAI}


One interesting question is why a model classifies a given image as some particular class, especially if the classification happens to be incorrect. Answering this question for deep neural networks, as we used them in our experiments, is a non-trivial problem, for which a great variety of different methods has been developed~\cite{Jeyakumar2020}. One of these methods, \emph{layer-wise relevance propagation} (LRP)~\cite{Montavon2019}, works particularly well for images, so we decided to use it for explaining which input features (i.\,e., image regions and -details) the trained models mainly base their decisions on. In a nutshell, LRP proceeds by first applying a model to some input (forward pass), and then propagating the activations from the top layer back to the input layer, layer by layer (backward pass). The details of this procedure are intricate; we refer the interested reader to~\cite{Montavon2019} for more information, only noting that among the multitude of possible parameter configurations of LRP we adhered to the one suggested for convolutional neural networks in~\cite{Kohlbrenner2020}.

In the remainder, we exclusively consider explanations for those models trained on all corruption intensities.

\begin{figure}
	\centering
	\subfloat[AT$_c$-AT$_c$]{
		\includegraphics[width=0.235\linewidth]{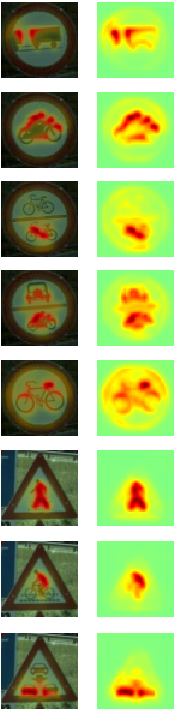}
	}
	\quad
	\subfloat[AT$_n$-AT$_n$]{
		\includegraphics[width=0.235\linewidth]{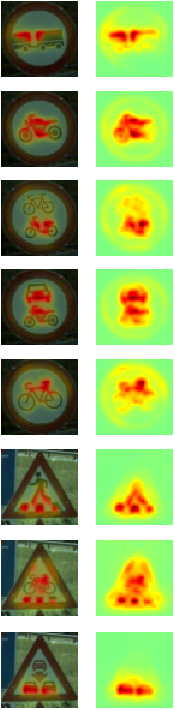}
	}
	\quad
	\subfloat[DE-DE]{
		\includegraphics[width=0.235\linewidth]{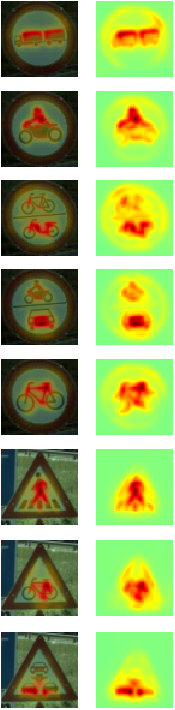}
	}
	\caption{Average explanations of all correctly predicted test images of some selected classes. Reddish to yellowish hues indicate regions with evidence \emph{in favor} of the predicted class, greenish hues indicate regions without any relevance, and bluish hues would indicate regions with evidence \emph{against} the predicted class (not present here).}
	\label{fig::correct_own_all_selected}
\end{figure}

Figure~\ref{fig::correct_own_all_selected} shows the average explanations of all correctly predicted test images of some selected classes, for the AT$_c$-AT$_c$, AT$_n$-AT$_n$ and DE-DE experiments. As usual in this field of research explanations are presented as heatmaps, where the color of a pixel indicates its relevance for the model. For each experiment, the left column blends the heatmaps with the actual images for the sake of the spatial context, whereas the right column only shows the heatmaps themselves. Explanations for the remaining classes can be found in Appendix~\ref{sec::Heatmaps}.

As can be seen, all models strongly focus on the pictograms (or parts of them) when classifying a traffic sign image, and only sometimes take the border of the sign into account as well. On the one hand this means that our models learned to pay attention to the `right' details of an image and do not base their decisions on spurious artifacts in the background, and on the other hand it means that the shape of the traffic signs does not really aid the models. This is not surprising, since the uniform circular and triangular shapes carry only little information for classifying the signs -- especially if the pictograms alone are sufficient for that purpose. Only in some cases, where the pictograms of prohibitory and warning signs are similar in appearance, taking the shape into account can be beneficial. One prominent example for this phenomenon are classes `\Fahrrad' and `\Fahrradueberfahrt': for the models trained on current Austrian pictograms the shape of `\Fahrrad' seems to be quite important, whereas the other models pay more attention to the shape of `\Fahrradueberfahrt'.

In classes `\Fussgaengeruebergang' and, to some extent, `\Fahrradueberfahrt', the models trained on proposed new Austrian and German pictograms focus a lot on the zebra crossing at the bottom. The models trained on the current Austrian pictograms, on the other hand, completely ignore the (only barely visible) zebra crossing and instead focus on the person. This difference in attention might be one of the reasons why in AT$_c$-AT$_n$ `\Fahrradueberfahrt' and `\Fussgaengeruebergang' only achieve a comparatively low accuracy of $28.93\%$ and $69.87\%$, respectively (cf. Table~\ref{tab::class_acc_foreign}).

It can also be observed that sometimes the models only look at certain parts of the pictograms, not the whole pictograms. This effect is particularly visible in class `\Falschfahrer', where all models completely ignore the car at the top and almost entirely ignore the arrow as well. Apparently, the two cars at the bottom are sufficient for robustly distinguishing this class from the other 23 traffic sign classes in our experiments. Likewise, in class `\FahrradMoped' the moped receives a lot more attention than the cycle, especially in AT$_c$-AT$_c$. Interestingly, in class `\KFZ', whose pictogram is similarly split into an upper and a lower part, the relevance is distributed much more evenly across the car and the motorcycle.

Classes `\AnhaengerLKW' and `\Motorrad' allow us to speculate why the models fail to generalize to other pictogram designs in some cases. Namely, both classes differ between the three design groups in certain aspects the models pay a lot attention to. The fact that the truck in `\AnhaengerLKW' is visible in its entirety in the German design seems to be important to the models, since quite some relevance is assigned to the front part of the truck. But also when comparing the two Austrian designs of this class one can observe a subtle difference in the relevance pattern: only a small amount of the truck is visible in the current Austrian design, leading to a vertical relevance pattern; in contrast, the proposed new Austrian design displays a slightly larger part of the truck, leading to a more horizontal pattern. Similarly, the fact that the current Austrian and German designs feature a person riding the motorcycle in class `\Motorrad' seems to be important to the models. The proposed new Austrian design lacks a rider, which the models seem to compensate by paying more attention to the front wheel.

It must be noted, though, that further experiments would be necessary to confirm the hypotheses expressed in the preceding paragraph. The relevance patterns constructed by LRP or any other feature attribution method are only meant to illustrate which parts of an image are important to a model, but care must be taken when trying to draw conclusions why the model fails to classify some class correctly.

\section{Discussion}
\label{sec::Discussion}

The objective of our work was to answer three research questions regarding the machine readability of traffic signs:
\begin{enumerate}
	\item Is there any significant difference between different pictogram design groups (current Austrian, proposed new Austrian, German) in terms of machine readability?
	\item Can traffic sign classification models trained on one pictogram design be safely deployed to traffic signs featuring a different design?
	\item Can general `design rules' for pictograms be formulated, for improving machine readability?
\end{enumerate}

For answering the first question we performed a two-sided $t$-test for testing the null-hypothesis of whether the average accuracies computed of the three independent runs, as reported in Table~\ref{tab::acc}, are equal. The $p$-values thus obtained are all greater than $0.09$, meaning that the null-hypothesis must be accepted under a usual confidence level of $95\%$. Hence, there is no statistically significant difference between the overall classification accuracies. This claim is also perfectly in line with, and hence further supported by, the results of the ResNet models (cf. Appendix~\ref{sec::ResNet}).


The answer to the second question is also negative: if any of the models trained on one pictogram design is applied to a different design, its classification accuracy drops significantly, by about $15\%$--$23\%$. Here, it is particularly interesting to note that a few classes cause massive problems, whereas most of the others can still be classified accurately. Even more surprising is the fact that our models consistently generalize best between German and \emph{current} Austrian pictograms (in both directions), although a human would probably find more similarities between German and \emph{new} Austrian pictograms. We do not have any explanation for this phenomenon.

Answering the third question is more intricate. Although we generated explanations for the models' predictions in Section~\ref{sec::XAI}, formulating design rules for pictograms based on them is difficult. Still, what \emph{can} be said is that machines (deep neural networks, in particular) perceive traffic signs differently than humans: humans try to \emph{understand} the meaning of pictograms in order to classify them, machines only try to \emph{distinguish} them. This is not surprising, as it is exactly what they are trained to do; distinguishing a fixed set of pictograms, however, might be possible based on small, semantically meaningless details, as discussed in Section~\ref{sec::XAI}. We hypothesize that this phenomenon is the main reason why the models in our experiment fail to generalize to `foreign' pictogram designs. Unfortunately, it is hardly possible to predict a-priori which details will be important to a classification model. The only general rule that can be formulated in this regard concerns the visibility of pictogram elements: the zebra crossings in classes `\Fahrradueberfahrt' and `\Fussgaengeruebergang' of the current Austrian designs consist of small, thin patches that quickly become imperceptible when the images are corrupted, and hence the models do not pay attention to them. In the proposed new Austrian design the zebra crossings are far more pronounced and thus better visible, and the models \emph{do} take them into account. Avoiding thin lines and overly small patches of ink has been one of the main design criteria for the new Austrian pictograms, and so can be expected to improve human readability of traffic signs as well.

Summarizing, the main takeaways of our work are as follows:
\begin{itemize}
	\item The driving factor of traffic sign pictogram design should remain human readability. Machines can handle different designs equally well, provided they have been trained on them.
	\item In the realistic scenario that a driving assistance system should correctly recognize traffics signs with different pictogram designs, as can be found in different countries, the models must be trained appropriately. This can be achieved by either training one single classifier on a data set encompassing many different designs, or by training a separate classifier for each design.
	\item If existing pictograms are replaced by a new design, classification models will likely have to be updated.
	Since acquiring large real-world data sets is time-consuming and only possible once the actual traffics signs have been replaced, it might be necessary to resort to synthetic data sets as presented in this paper instead.
\end{itemize}


\subsection{Limitations and Future Work}
\label{sec::FutureWork}

When defining our experimental setup we had to fix certain parameter values that are up to discussion and could be revised in future extensions of our experiments. First, due to the constraints mentioned in Section~\ref{sec::ClassSelection}, we only considered 24 traffic sign classes from categories `prohibitory' and `warning'. Actual classifiers deployed in driving assistance systems must be trained on a much wider variety of classes and may hence exhibit a different behavior \wrt\ sensitivity to pictogram design, frequently confused classes, and attention pattern. Still, we believe that our reduced setting approximates reality sufficiently well for making our findings hold more generally.

Another point of discussion concerns the image corruption strategy we employed. From the vast space of conceivable corruption methods we picked some that we deemed either realistic or particularly interesting, but it should be clear that many others would have been at our disposal, too. In future experiments one could in particular try to incorporate corruptions that are specific to traffic signs, like some kind of `over-exposure' where, due to the production process and reflectivity of the traffic sign foil, brighter areas seem to `grow' and hide parts of darker neighboring areas, making small and fine pictogram elements seem to disappear.
Furthermore, we focused on simulating distance by spatially downsampling the images at varying degrees, but we did not apply other geometric transformations like rotations and perspective distortions. In addition to the degree of downsampling, one could systematically vary the intensities of the `secondary' corruptions (rain, noise, blur, etc.) as well.

As discussed above, the models we obtained are not very robust \wrt\ `foreign' pictogram designs. One way to counter this could be forcing the models to pay more attention to the global shape of the pictograms, instead of small details. This, in turn, can perhaps be achieved by borrowing ideas from the current research on \emph{adversarial attacks}~\cite{Szegedy2014, Goodfellow2015}, like \emph{adversarial training}~\cite{Madry2019, Tack2021}. Adversarially trained models are usually more robust than normally trained ones, paying more attention to features that are also relevant for humans, but at the same time tend to perform worse on `clean' samples taken from the distribution they were trained on~\cite{Tsipras2019}. Alternatively one could also try to preprocess the images before training and applying a model, by applying a low-pass- or bilateral filter that destroys high-frequency information and thereby biases the model towards low-frequency shape information. Repeating our experiments with adversarially trained models and/or said input preprocessing would be an interesting direction for future research.

Finally, we want to conclude this paper with some sort of `disclaimer'. State of the art driving assistance systems are black boxes, meaning that we do not know which traffic sign recognition mechanisms they employ. These mechanisms might not even resort to machine learning techniques, but could in principle be based on classical image processing methods that rely on hand-crafted features and representations instead of learned ones. The huge recent success of artificial neural networks in this application domain renders such a scenario unrealistic, though -- but even if neural networks are used, they can be implemented and trained in many different ways (regarding network architecture, ensembles of models, etc.). We want to encourage the developers of driving assistance systems to take the results of our experiments into account and, if appropriate, conduct similar experiments with their own traffic sign classifiers.

\paragraph*{Acknowledgments.}
We thank Stefan Egger for providing us with the proposed new Austrian pictogram designs, for his valuable suggestions regarding our experimental setup, and for his comments on the manuscript.

This research was funded by FFG (Austrian Research Promotion Agency) under grant 879320 (SafeSign) and supported by the strategic economic research programme ``Innovatives O\"O 2020'' of the province of Upper Austria.

\printbibliography

\appendix
\section{List of Traffic Sign Classes and Pictograms}
\label{sec::TSList}

\begin{longtable}{p{0.15\linewidth}p{0.25\linewidth}|ccc}
Name & Description & AT$_c$ & AT$_n$ & DE\\\hline\hline

\Achslast & No vehicles having a weight exceeding $n$ tonnes on one axle & \raisebox{-0.8\totalheight}{\includegraphics[width=0.1\linewidth]{img/sign_classes/old/Achslast.png}} &
\raisebox{-0.8\totalheight}{\includegraphics[width=0.1\linewidth]{img/sign_classes/new/Achslast.png}} &
\raisebox{-0.8\totalheight}{\includegraphics[width=0.1\linewidth]{img/sign_classes/DE/Achslast.png}}\\\hline

\Anhaenger* & No power-driven vehicles drawing a trailer & \raisebox{-0.8\totalheight}{\includegraphics[width=0.1\linewidth]{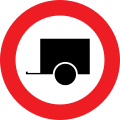}} &
\raisebox{-0.8\totalheight}{\includegraphics[width=0.1\linewidth]{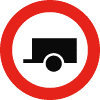}} &
\raisebox{-0.8\totalheight}{\includegraphics[width=0.1\linewidth]{img/sign_classes/DE/Anhaenger.png}}\\\hline

\AnhaengerLKW & No trucks drawing a trailer & \raisebox{-0.8\totalheight}{\includegraphics[width=0.1\linewidth]{img/sign_classes/old/Anhaenger_LKW.png}} &
\raisebox{-0.8\totalheight}{\includegraphics[width=0.1\linewidth]{img/sign_classes/new/Anhaenger_LKW.png}} &
\raisebox{-0.8\totalheight}{\includegraphics[width=0.1\linewidth]{img/sign_classes/DE/Anhaenger_LKW.png}}\\\hline

\LKW & No trucks & \raisebox{-0.8\totalheight}{\includegraphics[width=0.1\linewidth]{img/sign_classes/old/LKW.png}} &
\raisebox{-0.8\totalheight}{\includegraphics[width=0.1\linewidth]{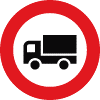}} &
\raisebox{-0.8\totalheight}{\includegraphics[width=0.1\linewidth]{img/sign_classes/DE/LKW.png}}\\\hline

\GewichtLKW* & No trucks having a weight exceeding $n$ tonnes & \raisebox{-0.8\totalheight}{\includegraphics[width=0.1\linewidth]{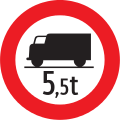}} &
\raisebox{-0.8\totalheight}{\includegraphics[width=0.1\linewidth]{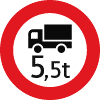}} &
\raisebox{-0.8\totalheight}{\includegraphics[width=0.1\linewidth]{img/sign_classes/DE/Gewicht_LKW.png}}\\\hline

\Omnibus & No omnibuses & \raisebox{-0.8\totalheight}{\includegraphics[width=0.1\linewidth]{img/sign_classes/old/Omnibus.png}} &
\raisebox{-0.8\totalheight}{\includegraphics[width=0.1\linewidth]{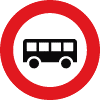}} &
\raisebox{-0.8\totalheight}{\includegraphics[width=0.1\linewidth]{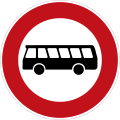}}\\\hline

\Motorrad & No single-tracked motorcycles & \raisebox{-0.8\totalheight}{\includegraphics[width=0.1\linewidth]{img/sign_classes/old/Motorrad.png}} &
\raisebox{-0.8\totalheight}{\includegraphics[width=0.1\linewidth]{img/sign_classes/new/Motorrad.png}} &
\raisebox{-0.8\totalheight}{\includegraphics[width=0.1\linewidth]{img/sign_classes/DE/Motorrad.png}}\\\hline

\Moped & No mopeds & \raisebox{-0.8\totalheight}{\includegraphics[width=0.1\linewidth]{img/sign_classes/old/Moped.png}} &
\raisebox{-0.8\totalheight}{\includegraphics[width=0.1\linewidth]{img/sign_classes/new/Moped.png}} &
\raisebox{-0.8\totalheight}{\includegraphics[width=0.1\linewidth]{img/sign_classes/DE/Moped.png}}\\\hline

\Fahrrad & No cycles & \raisebox{-0.8\totalheight}{\includegraphics[width=0.1\linewidth]{img/sign_classes/old/Fahrrad.png}} &
\raisebox{-0.8\totalheight}{\includegraphics[width=0.1\linewidth]{img/sign_classes/new/Fahrrad.png}} &
\raisebox{-0.8\totalheight}{\includegraphics[width=0.1\linewidth]{img/sign_classes/DE/Fahrrad.png}}\\\hline

\FahrradMoped & No cycles or mopeds & \raisebox{-0.8\totalheight}{\includegraphics[width=0.1\linewidth]{img/sign_classes/old/Fahrrad_Moped.png}} &
\raisebox{-0.8\totalheight}{\includegraphics[width=0.1\linewidth]{img/sign_classes/new/Fahrrad_Moped.png}} &
\raisebox{-0.8\totalheight}{\includegraphics[width=0.1\linewidth]{img/sign_classes/DE/Fahrrad_Moped.png}}\\\hline

\KFZ & No power-driven vehicles & \raisebox{-0.8\totalheight}{\includegraphics[width=0.1\linewidth]{img/sign_classes/old/KFZ.png}} &
\raisebox{-0.8\totalheight}{\includegraphics[width=0.1\linewidth]{img/sign_classes/new/KFZ.png}} &
\raisebox{-0.8\totalheight}{\includegraphics[width=0.1\linewidth]{img/sign_classes/DE/KFZ.png}}\\\hline

\Einspurige & No power-driven vehicles except single-tracked motorcycles & \raisebox{-0.8\totalheight}{\includegraphics[width=0.1\linewidth]{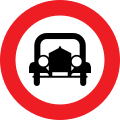}} &
\raisebox{-0.8\totalheight}{\includegraphics[width=0.1\linewidth]{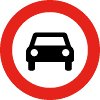}} &
\raisebox{-0.8\totalheight}{\includegraphics[width=0.1\linewidth]{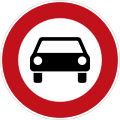}}\\\hline

\Reitverbot & Riding prohibited & \raisebox{-0.8\totalheight}{\includegraphics[width=0.1\linewidth]{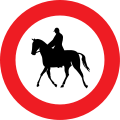}} &
\raisebox{-0.8\totalheight}{\includegraphics[width=0.1\linewidth]{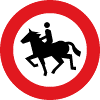}} &
\raisebox{-0.8\totalheight}{\includegraphics[width=0.1\linewidth]{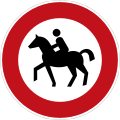}}\\\hline

\Fuhrwerk & No animal-drawn vehicles & \raisebox{-0.8\totalheight}{\includegraphics[width=0.1\linewidth]{img/sign_classes/old/Fuhrwerk.png}} &
\raisebox{-0.8\totalheight}{\includegraphics[width=0.1\linewidth]{img/sign_classes/new/Fuhrwerk.png}} &
\raisebox{-0.8\totalheight}{\includegraphics[width=0.1\linewidth]{img/sign_classes/DE/Fuhrwerk.png}}\\\hline

\Ueberholverbot & No overtaking & \raisebox{-0.8\totalheight}{\includegraphics[width=0.1\linewidth]{img/sign_classes/old/Ueberholverbot.png}} &
\raisebox{-0.8\totalheight}{\includegraphics[width=0.1\linewidth]{img/sign_classes/new/Ueberholverbot.png}} &
\raisebox{-0.8\totalheight}{\includegraphics[width=0.1\linewidth]{img/sign_classes/DE/Ueberholverbot.png}}\\\hline

\UeberholverbotLKW & No overtaking by trucks & \raisebox{-0.8\totalheight}{\includegraphics[width=0.1\linewidth]{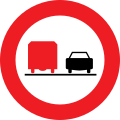}} &
\raisebox{-0.8\totalheight}{\includegraphics[width=0.1\linewidth]{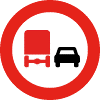}} &
\raisebox{-0.8\totalheight}{\includegraphics[width=0.1\linewidth]{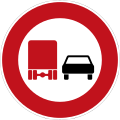}}\\\hline

\Gefahrengut & No vehicles carrying dangerous goods & \raisebox{-0.8\totalheight}{\includegraphics[width=0.1\linewidth]{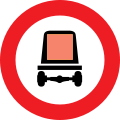}} &
\raisebox{-0.8\totalheight}{\includegraphics[width=0.1\linewidth]{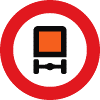}} &
\raisebox{-0.8\totalheight}{\includegraphics[width=0.1\linewidth]{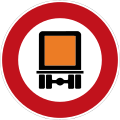}}\\\hline

\Fussgaenger & Pedestrians prohibited & \raisebox{-0.8\totalheight}{\includegraphics[width=0.1\linewidth]{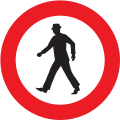}} &
\raisebox{-0.8\totalheight}{\includegraphics[width=0.1\linewidth]{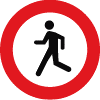}} &
\raisebox{-0.8\totalheight}{\includegraphics[width=0.1\linewidth]{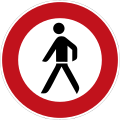}}\\\hline

\Baustelle & Warning: road works & \raisebox{-0.8\totalheight}{\includegraphics[width=0.1\linewidth]{img/sign_classes/old/Baustelle.png}} &
\raisebox{-0.8\totalheight}{\includegraphics[width=0.1\linewidth]{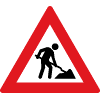}} &
\raisebox{-0.8\totalheight}{\includegraphics[width=0.1\linewidth]{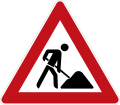}}\\\hline

\Kinder & Warning: children & \raisebox{-0.8\totalheight}{\includegraphics[width=0.1\linewidth]{img/sign_classes/old/Kinder.png}} &
\raisebox{-0.8\totalheight}{\includegraphics[width=0.1\linewidth]{img/sign_classes/new/Kinder.png}} &
\raisebox{-0.8\totalheight}{\includegraphics[width=0.1\linewidth]{img/sign_classes/DE/Kinder.png}}\\\hline

\Fussgaengeruebergang & Warning: pedestrian crossing & \raisebox{-0.8\totalheight}{\includegraphics[width=0.1\linewidth]{img/sign_classes/old/Fussgaengeruebergang.png}} &
\raisebox{-0.8\totalheight}{\includegraphics[width=0.1\linewidth]{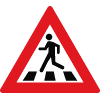}} &
\raisebox{-0.8\totalheight}{\includegraphics[width=0.1\linewidth]{img/sign_classes/DE/Fussgaengeruebergang.png}}\\\hline

\Fahrradueberfahrt* & Warning: cyclist crossing & \raisebox{-0.8\totalheight}{\includegraphics[width=0.1\linewidth]{img/sign_classes/old/Fahrradueberfahrt.png}} &
\raisebox{-0.8\totalheight}{\includegraphics[width=0.1\linewidth]{img/sign_classes/new/Fahrradueberfahrt.png}} &
\raisebox{-0.8\totalheight}{\includegraphics[width=0.1\linewidth]{img/sign_classes/DE/Fahrradueberfahrt.png}}\\\hline

\Schleudergefahr & Warning: slippery road & \raisebox{-0.8\totalheight}{\includegraphics[width=0.1\linewidth]{img/sign_classes/old/Schleudergefahr.png}} &
\raisebox{-0.8\totalheight}{\includegraphics[width=0.1\linewidth]{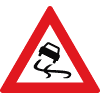}} &
\raisebox{-0.8\totalheight}{\includegraphics[width=0.1\linewidth]{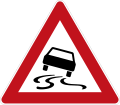}}\\\hline

\Falschfahrer* & Warning: motorist driving against the traffic on motorways  & \raisebox{-0.8\totalheight}{\includegraphics[width=0.1\linewidth]{img/sign_classes/old/Falschfahrer.png}} &
\raisebox{-0.8\totalheight}{\includegraphics[width=0.1\linewidth]{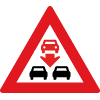}} &
\raisebox{-0.8\totalheight}{\includegraphics[width=0.1\linewidth]{img/sign_classes/DE/Falschfahrer.png}}\\

\caption{List of all traffic sign classes used in our experiments. Signs marked with * exist in Austria but do not have German counterparts, meaning that we hand-crafted them for our experiments.\label{tab::ClassList}}
\end{longtable}

Table~\ref{tab::ClassList} lists all traffic sign classes and pictograms we used in our experiments. The way how we obtained the German pictograms for those classes not existing in Germany is described in Section~\ref{sec::ClassSelection}.

\section{Detailed Per-Class Results}
\label{sec::DetailedClass}

\begin{table}
	\centering
	\small
	\begin{tabular}{l|rrrrrrr}
		                    &  All &   Per-level rank &     1 &    2 &    3 &    4 &    5 \\\hline
		\Gefahrengut          & 99.67 &   5.9 ($\pm$7.3) & 100.00 &  99.67 & 99.33 & 98.67 & 96.67 \\
		\Achslast             & 99.67 &   5.1 ($\pm$9.3) &  99.67 & 100.00 & 99.33 & 99.67 & 97.33 \\
		\Baustelle            & 99.60 &   8.9 ($\pm$6.7) &  99.33 &  99.67 & 99.67 & 99.00 & 95.33 \\
		\Reitverbot           & 99.47 &   6.2 ($\pm$7.2) & 100.00 & 100.00 & 99.00 & 98.00 & 96.67 \\
		\FahrradMoped         & 99.40 &  14.8 ($\pm$3.9) &  99.67 &  99.33 & 98.00 & 97.00 & 95.00 \\
		\KFZ                  & 99.40 &  10.7 ($\pm$5.4) & 100.00 & 100.00 & 97.67 & 98.00 & 93.67 \\
		\Anhaenger            & 99.33 &   7.2 ($\pm$7.2) & 100.00 &  99.67 & 99.67 & 99.33 & 94.00 \\
		\Einspurige           & 99.27 &   9.7 ($\pm$4.1) & 100.00 &  99.33 & 98.67 & 98.33 & 96.00 \\
		\Fahrrad              & 99.27 &  12.7 ($\pm$5.2) &  99.00 &  99.67 & 97.67 & 97.67 & 96.33 \\
		\GewichtLKW           & 99.20 &   6.8 ($\pm$7.6) & 100.00 & 100.00 & 99.67 & 97.33 & 96.33 \\
		\Fussgaenger          & 99.13 &   9.9 ($\pm$6.0) & 100.00 &  99.00 & 98.00 & 98.67 & 96.33 \\
		\Fuhrwerk             & 99.07 &   8.3 ($\pm$5.8) & 100.00 & 100.00 & 97.67 & 98.33 & 96.00 \\
		\AnhaengerLKW         & 99.07 &   9.0 ($\pm$5.0) & 100.00 &  99.67 & 97.67 & 99.00 & 95.67 \\
		\Kinder               & 99.00 &  14.6 ($\pm$4.3) &  99.00 &  99.33 & 97.67 & 98.00 & 95.67 \\
		\UeberholverbotLKW    & 98.80 &  10.7 ($\pm$4.9) & 100.00 &  99.67 & 97.33 & 97.67 & 96.33 \\
		\Motorrad             & 98.80 &  17.3 ($\pm$5.2) &  99.33 &  99.33 & 97.33 & 97.00 & 94.00 \\
		\Schleudergefahr      & 98.53 &  13.7 ($\pm$5.9) & 100.00 &  99.33 & 99.00 & 97.33 & 92.67 \\
		\Falschfahrer         & 98.53 &  11.3 ($\pm$6.7) &  99.67 &  99.67 & 99.67 & 98.33 & 92.67 \\
		\LKW                  & 98.47 &  15.8 ($\pm$6.4) & 100.00 &  99.33 & 96.00 & 97.00 & 94.67 \\
		\Moped                & 98.33 &  18.7 ($\pm$7.6) &  99.00 &  98.67 & 98.00 & 97.33 & 89.67 \\
		\Omnibus              & 98.13 &  17.0 ($\pm$7.2) & 100.00 &  99.33 & 96.67 & 95.00 & 93.67 \\
		\Fahrradueberfahrt    & 98.07 &  22.6 ($\pm$8.9) &  98.33 &  98.33 & 95.67 & 96.33 & 92.67 \\
		\Ueberholverbot       & 97.87 &  20.8 ($\pm$7.3) &  99.00 &  99.00 & 96.67 & 96.00 & 93.33 \\
		\Fussgaengeruebergang & 97.40 &  22.3 ($\pm$8.5) &  98.00 &  97.67 & 97.00 & 93.67 & 93.33
	\end{tabular}
	\caption{AT$_c$-AT$_c$ per-class accuracy (\%), sorted descending by the accuracy in the models trained on the whole data set (column `All'). Per-level rank is the average ($\pm$ standard deviation) rank of each class for the models trained separately on the five corruption intensities (columns 1--5).}
	\label{tab::old_classes}
\end{table}

\begin{table}
	\centering
	\small
	\begin{tabular}{l|rrrrrrr}
							&  All &   Per-level rank &     1 &    2 &    3 &    4 &    5 \\\hline
		\Fussgaenger          & 99.53 &  6.7 ($\pm$10.1) & 100.00 &  99.00 & 100.00 & 99.00 & 96.67 \\
		\Fussgaengeruebergang & 99.27 &  11.0 ($\pm$7.1) &  99.33 & 100.00 & 100.00 & 96.67 & 95.00 \\
		\Fuhrwerk             & 99.27 &  12.9 ($\pm$6.2) &  99.67 &  99.33 &  98.33 & 97.00 & 97.00 \\
		\Falschfahrer         & 99.13 &   9.3 ($\pm$4.9) &  99.67 & 100.00 &  99.00 & 97.67 & 95.67 \\
		\Achslast             & 99.13 &  10.5 ($\pm$4.6) &  99.67 &  99.33 &  99.33 & 98.00 & 95.67 \\
		\Gefahrengut          & 99.13 &   6.2 ($\pm$7.3) & 100.00 &  99.67 &  99.33 & 98.67 & 96.33 \\
		\Schleudergefahr      & 99.00 &   7.7 ($\pm$7.8) & 100.00 &  99.33 &  99.33 & 99.67 & 95.67 \\
		\Reitverbot           & 99.00 &  11.3 ($\pm$5.4) & 100.00 &  99.67 &  99.00 & 97.67 & 93.67 \\
		\Einspurige           & 98.93 &   9.9 ($\pm$5.9) &  99.67 & 100.00 &  99.33 & 98.00 & 94.00 \\
		\GewichtLKW           & 98.80 &   8.6 ($\pm$7.2) & 100.00 & 100.00 &  98.33 & 97.33 & 96.00 \\
		\FahrradMoped         & 98.80 &  10.9 ($\pm$5.1) &  99.67 &  99.67 &  98.33 & 98.67 & 95.67 \\
		\Kinder               & 98.73 &  11.4 ($\pm$4.4) &  99.67 &  99.67 &  99.67 & 97.00 & 94.33 \\
		\Baustelle            & 98.67 &   5.7 ($\pm$8.2) & 100.00 & 100.00 &  99.33 & 97.33 & 97.00 \\
		\Anhaenger            & 98.67 &  13.7 ($\pm$7.3) &  98.67 & 100.00 &  99.33 & 96.67 & 94.33 \\
		\Fahrradueberfahrt    & 98.60 &  17.0 ($\pm$5.8) &  99.33 &  99.33 &  99.00 & 95.67 & 95.00 \\
		\UeberholverbotLKW    & 98.60 &   8.1 ($\pm$6.4) &  99.67 &  99.67 &  99.67 & 99.00 & 95.33 \\
		\LKW                  & 98.60 &  14.3 ($\pm$5.7) &  99.67 &  99.33 &  99.33 & 94.67 & 95.33 \\
		\Fahrrad              & 98.60 &  13.6 ($\pm$4.5) &  99.67 &  99.33 &  98.33 & 97.33 & 95.67 \\
		\AnhaengerLKW         & 98.60 &  17.9 ($\pm$6.3) &  99.00 &  99.33 &  99.00 & 96.33 & 94.00 \\
		\KFZ                  & 98.20 &  19.4 ($\pm$8.1) &  99.33 &  99.67 &  98.00 & 93.67 & 90.67 \\
		\Ueberholverbot       & 98.13 &  15.0 ($\pm$4.0) &  99.67 &  99.33 &  98.67 & 97.00 & 94.67 \\
		\Omnibus              & 98.00 &  13.5 ($\pm$7.9) & 100.00 & 100.00 &  98.33 & 96.00 & 92.00 \\
		\Moped                & 97.87 &  22.5 ($\pm$8.5) &  98.67 &  99.00 &  97.67 & 96.00 & 89.33 \\
		\Motorrad             & 97.00 &  22.9 ($\pm$9.1) &  99.33 &  98.00 &  97.33 & 92.00 & 90.33
	\end{tabular}
	\caption{AT$_n$-AT$_n$ per-class accuracy (\%), sorted descending by the accuracy in the models trained on the whole data set (column `All'). Per-level rank is the average ($\pm$ standard deviation) rank of each class for the models trained separately on the five corruption intensities (columns 1--5).}
	\label{tab::new_classes}
\end{table}

\begin{table}
	\centering
	\small
	\begin{tabular}{l|rrrrrrr}
							&  All &   Per-level rank &     1 &    2 &    3 &    4 &    5 \\\hline
		\Baustelle            & 99.47 &  12.3 ($\pm$7.9) &  99.00 &  99.33 &  98.00 & 99.67 & 97.00 \\
		\GewichtLKW           & 99.40 &   5.3 ($\pm$8.0) & 100.00 &  99.67 &  99.67 & 98.33 & 98.67 \\
		\UeberholverbotLKW    & 99.40 &   6.6 ($\pm$6.3) & 100.00 & 100.00 &  99.67 & 98.00 & 95.67 \\
		\Achslast             & 99.40 &  10.3 ($\pm$5.4) &  99.67 &  99.67 &  99.33 & 97.00 & 97.67 \\
		\Fussgaenger          & 99.33 &   6.6 ($\pm$6.8) & 100.00 &  99.67 &  99.00 & 99.67 & 97.33 \\
		\Reitverbot           & 99.27 &   6.9 ($\pm$5.9) & 100.00 & 100.00 &  99.67 & 98.00 & 95.33 \\
		\Anhaenger            & 99.27 &   5.7 ($\pm$7.6) & 100.00 & 100.00 & 100.00 & 98.00 & 96.00 \\
		\Fussgaengeruebergang & 99.20 &  14.2 ($\pm$4.2) &  99.33 &  99.33 &  99.33 & 97.33 & 95.00 \\
		\Gefahrengut          & 99.13 &   5.9 ($\pm$7.1) & 100.00 &  99.67 &  99.33 & 98.67 & 98.33 \\
		\Motorrad             & 99.07 &  18.1 ($\pm$7.2) &  99.67 &  98.67 &  97.33 & 97.67 & 93.00 \\
		\FahrradMoped         & 99.07 &  12.7 ($\pm$6.1) &  99.67 &  99.33 &  97.67 & 98.33 & 96.67 \\
		\Einspurige           & 99.00 &   9.2 ($\pm$4.2) & 100.00 &  99.67 &  99.00 & 97.67 & 96.33 \\
		\Fuhrwerk             & 99.00 &  12.5 ($\pm$4.3) & 100.00 &  99.67 &  98.00 & 96.67 & 95.00 \\
		\LKW                  & 98.93 &  11.5 ($\pm$6.3) & 100.00 &  99.33 & 100.00 & 95.67 & 94.33 \\
		\Fahrradueberfahrt    & 98.67 &  14.2 ($\pm$4.6) & 100.00 &  99.00 &  98.67 & 97.33 & 94.00 \\
		\Schleudergefahr      & 98.67 &  16.8 ($\pm$8.3) &  99.00 &  98.33 &  99.33 & 95.33 & 96.33 \\
		\Moped                & 98.67 &  16.1 ($\pm$6.9) &  98.67 &  99.67 &  98.00 & 97.67 & 92.33 \\
		\Falschfahrer         & 98.60 &  10.2 ($\pm$8.6) & 100.00 &  99.33 & 100.00 & 98.67 & 91.00 \\
		\AnhaengerLKW         & 98.33 &  18.8 ($\pm$7.4) &  99.67 &  98.33 &  98.67 & 94.67 & 94.00 \\
		\Omnibus              & 98.33 &  13.6 ($\pm$7.7) & 100.00 & 100.00 &  98.33 & 95.33 & 92.00 \\
		\KFZ                  & 98.27 &  17.7 ($\pm$7.0) &  99.33 &  99.00 &  99.33 & 96.00 & 91.33 \\
		\Fahrrad              & 98.20 &  19.7 ($\pm$7.7) &  99.67 &  99.00 &  97.67 & 95.67 & 91.33 \\
		\Kinder               & 98.13 &  19.7 ($\pm$7.8) &  99.67 &  99.00 &  97.00 & 95.67 & 92.33 \\
		\Ueberholverbot       & 97.53 &  15.4 ($\pm$7.0) & 100.00 &  99.33 &  97.67 & 94.00 & 95.67
	\end{tabular}
	\caption{DE-DE per-class accuracy (\%), sorted descending by the accuracy in the models trained on the whole data set (column `All'). Per-level rank is the average ($\pm$ standard deviation) rank of each class for the models trained separately on the five corruption intensities (columns 1--5).}
	\label{tab::DE_classes}
\end{table}

Tables~\ref{tab::old_classes}, \ref{tab::new_classes} and~\ref{tab::DE_classes} list the accuracy of every individual class in the AT$_c$-AT$_c$, AT$_n$-AT$_n$ and DE-DE evaluations, respectively.

\begin{figure}
	\centering
	\includegraphics[width=\linewidth]{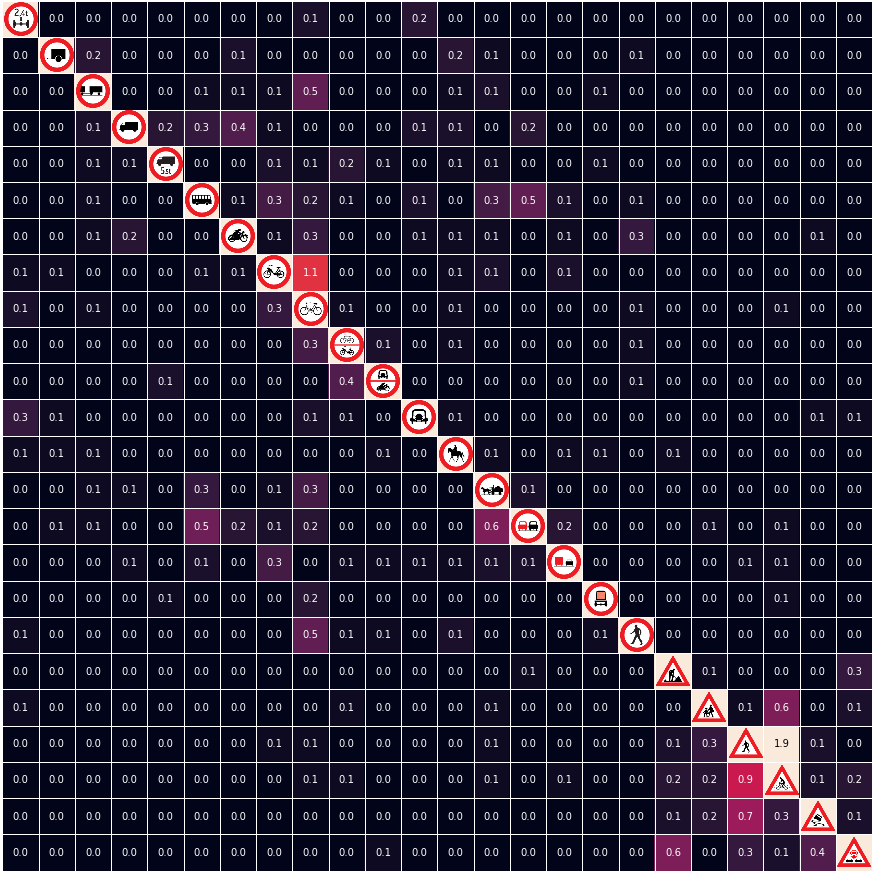}
	\caption{AT$_c$-AT$_c$ confusion matrix for the models trained on the whole data set. Rows represent true classes, columns represent predictions, numbers are percentages of misclassified samples (sum up to 100 row-wise).}
	\label{fig::confusion_old_all}
\end{figure}

\begin{figure}
	\centering
	\includegraphics[width=\linewidth]{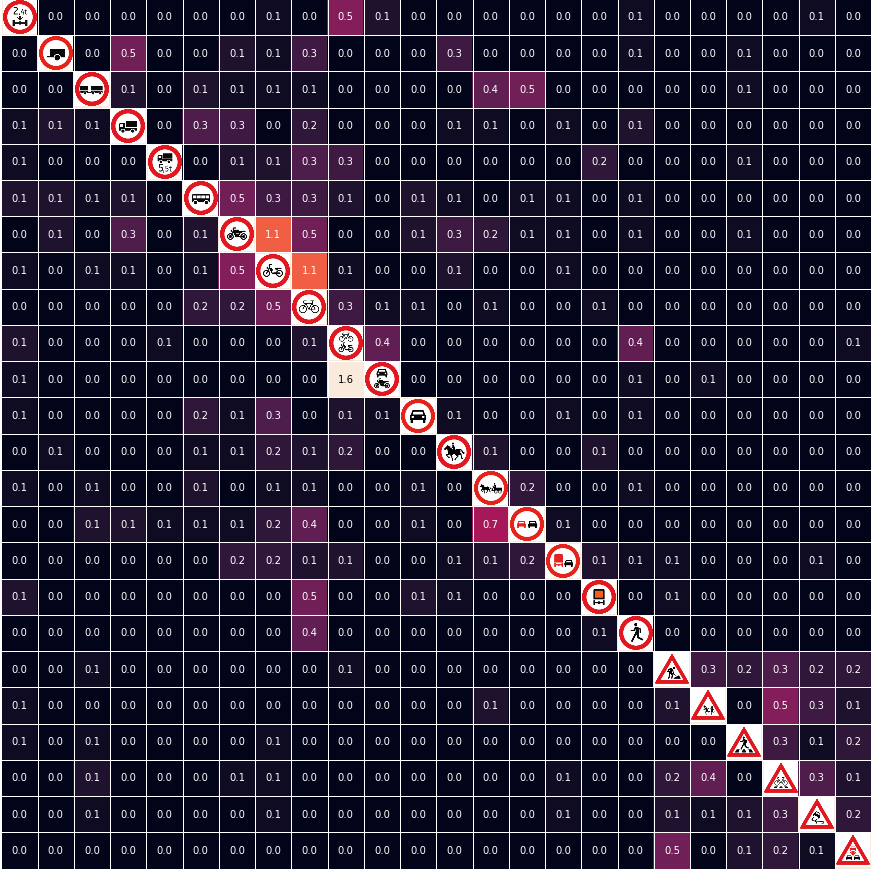}
	\caption{AT$_n$-AT$_n$ confusion matrix for the models trained on the whole data set. Rows represent true classes, columns represent predictions, numbers are percentages of misclassified samples (sum up to 100 row-wise).}
	\label{fig::confusion_new_all}
\end{figure}

\begin{figure}
	\centering
	\includegraphics[width=\linewidth]{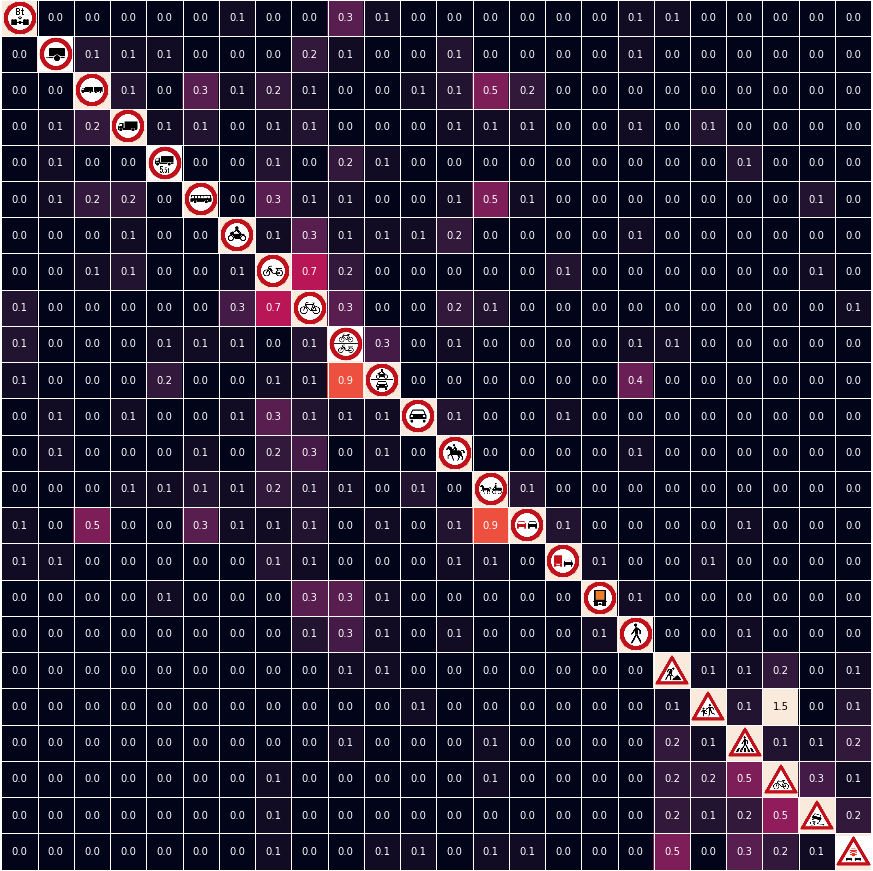}
	\caption{DE-DE confusion matrix for the models trained on the whole data set. Rows represent true classes, columns represent predictions, numbers are percentages of misclassified samples (sum up to 100 row-wise).}
	\label{fig::confusion_DE_all}
\end{figure}

Figures~\ref{fig::confusion_old_all}, \ref{fig::confusion_new_all} and~\ref{fig::confusion_DE_all} show the confusion matrices for AT$_c$-AT$_c$, AT$_n$-AT$_n$ and DE-DE, respectively.

\begin{figure}
	\centering
	\includegraphics[width=\linewidth]{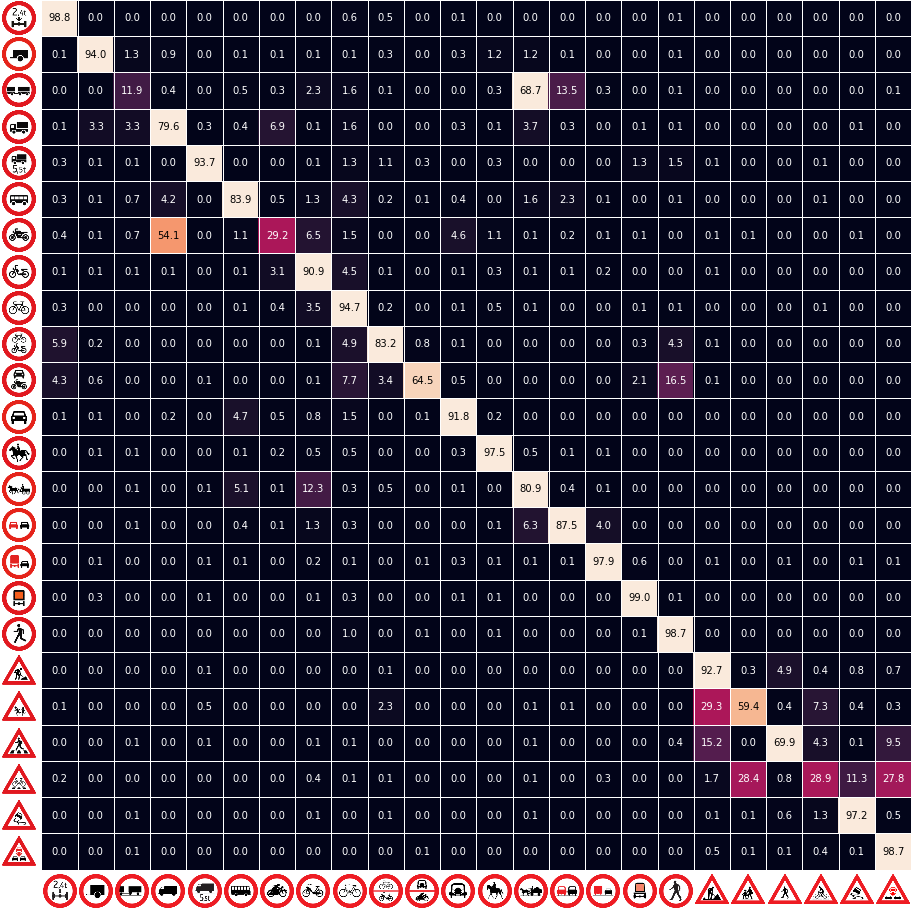}
	\caption{AT$_c$-AT$_n$ confusion matrix for the models trained on the whole data set. Rows represent true classes, columns represent predictions, numbers are percentages of misclassified samples (sum up to 100 row-wise).}
	\label{fig::confusion_old_new_all}
\end{figure}

\begin{figure}
	\centering
	\includegraphics[width=\linewidth]{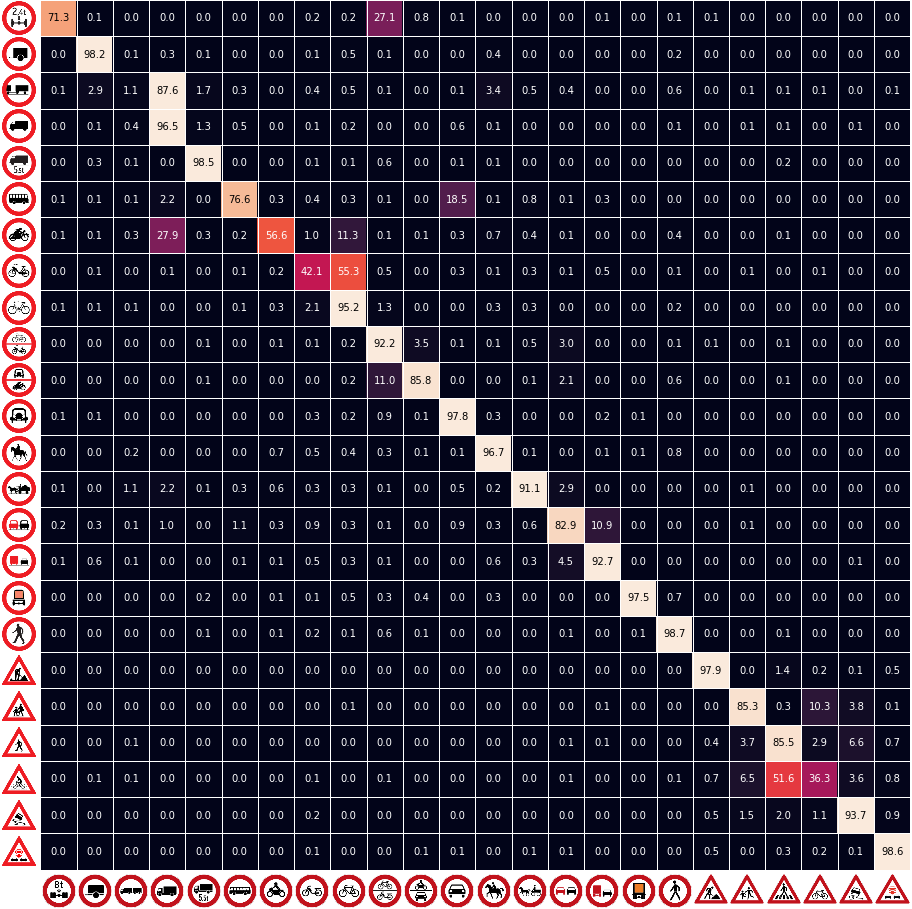}
	\caption{DE-AT$_c$ confusion matrix for the models trained on the whole data set. Rows represent true classes, columns represent predictions, numbers are percentages of misclassified samples (sum up to 100 row-wise).}
	\label{fig::confusion_DE_old_all}
\end{figure}

\begin{figure}
	\centering
	\includegraphics[width=\linewidth]{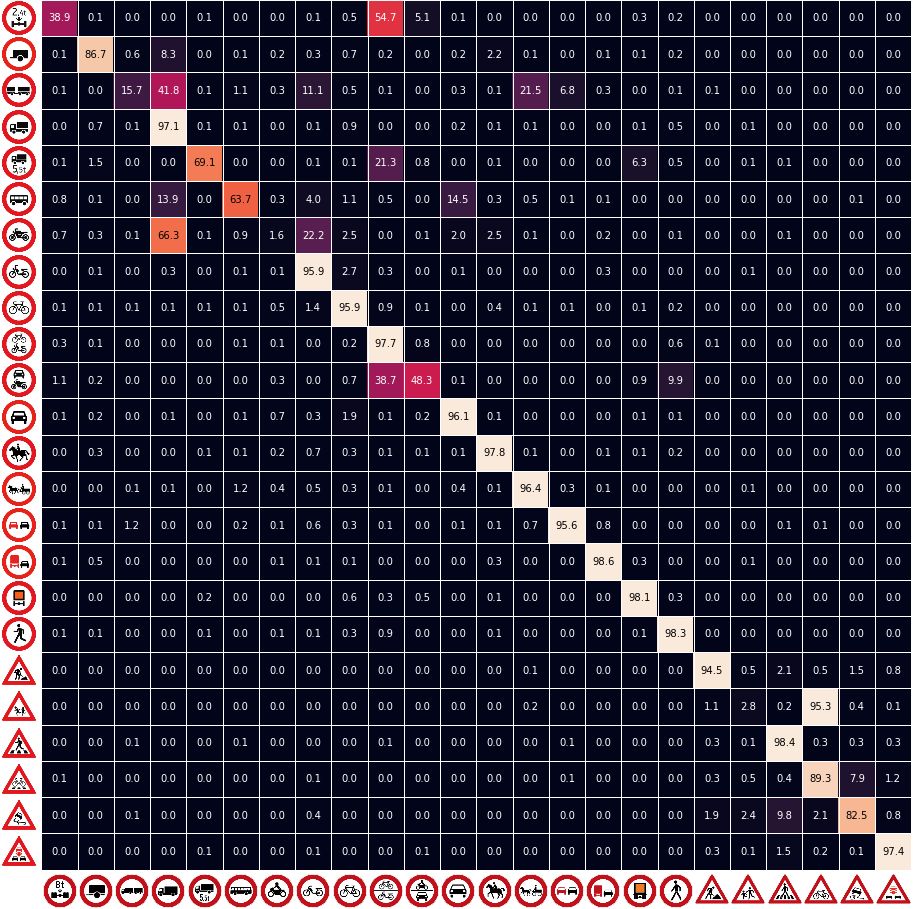}
	\caption{DE-AT$_n$ confusion matrix for the models trained on the whole data set. Rows represent true classes, columns represent predictions, numbers are percentages of misclassified samples (sum up to 100 row-wise).}
	\label{fig::confusion_DE_new_all}
\end{figure}

\begin{figure}
	\centering
	\includegraphics[width=\linewidth]{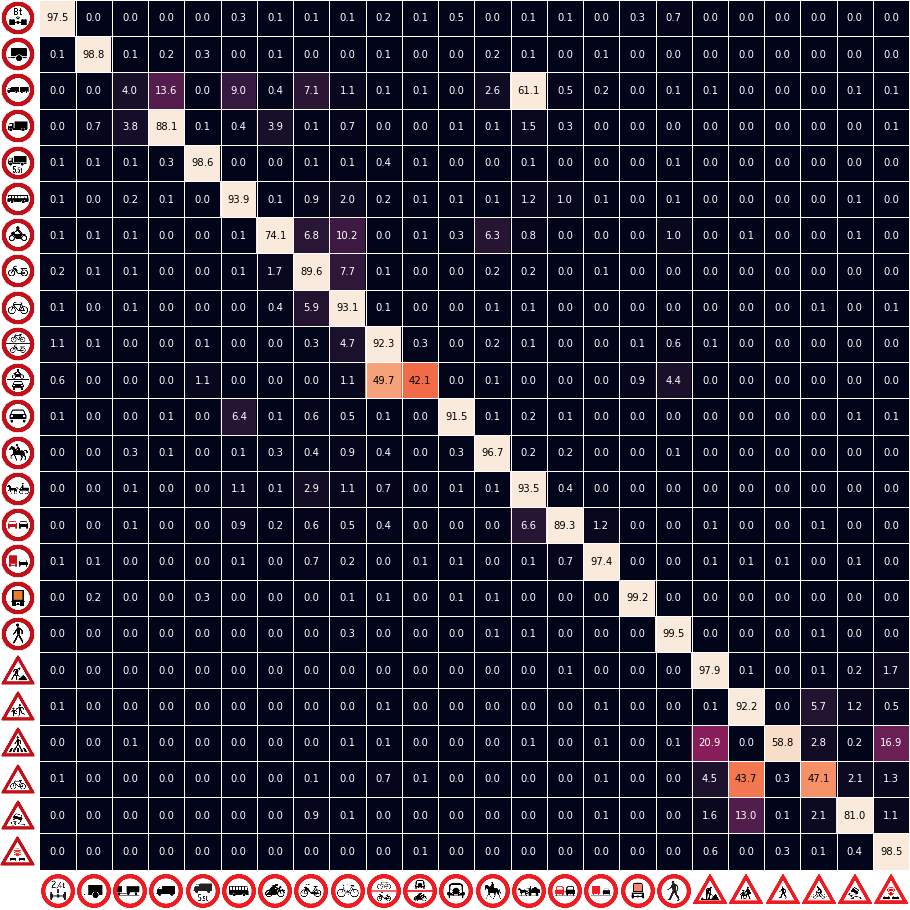}
	\caption{AT$_c$-DE confusion matrix for the models trained on the whole data set. Rows represent true classes, columns represent predictions, numbers are percentages of misclassified samples (sum up to 100 row-wise).}
	\label{fig::confusion_old_DE_all}
\end{figure}

\begin{figure}
	\centering
	\includegraphics[width=\linewidth]{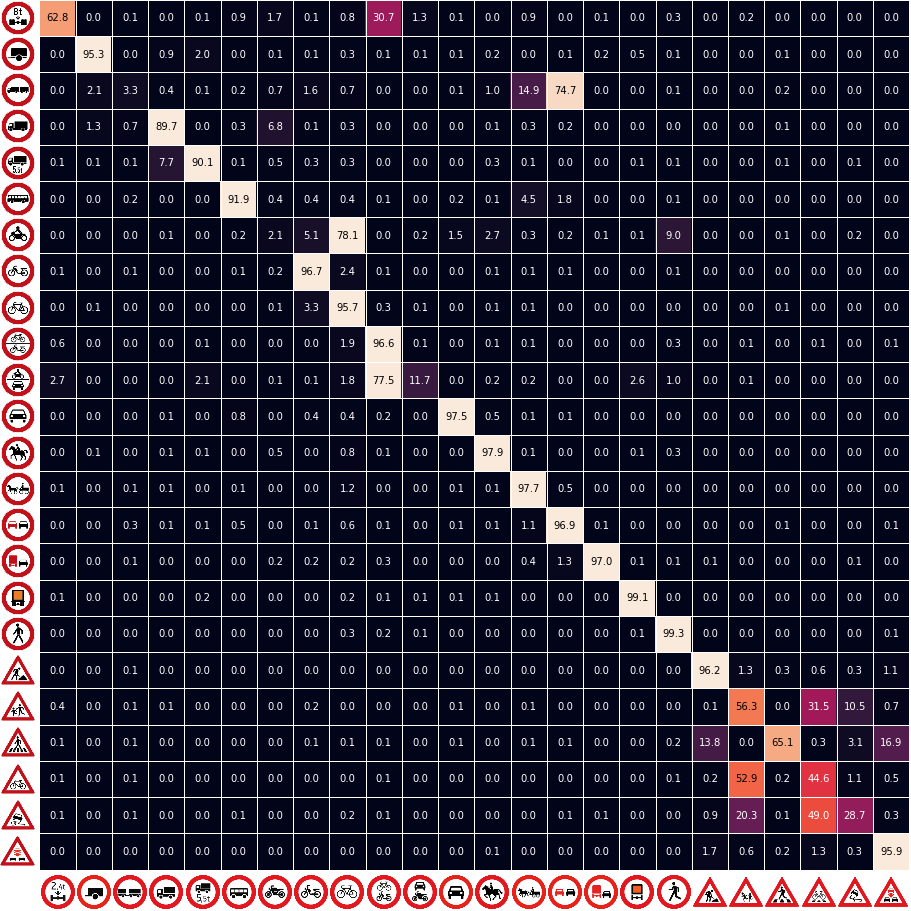}
	\caption{AT$_n$-DE confusion matrix for the models trained on the whole data set. Rows represent true classes, columns represent predictions, numbers are percentages of misclassified samples (sum up to 100 row-wise).}
	\label{fig::confusion_new_DE_all}
\end{figure}

Figures~\ref{fig::confusion_old_new_all}, \ref{fig::confusion_DE_old_all}, \ref{fig::confusion_DE_new_all}, \ref{fig::confusion_old_DE_all} and~\ref{fig::confusion_new_DE_all} show the confusion matrices for AT$_c$-AT$_n$, DE-AT$_c$, DE-AT$_n$, AT$_c$-DE and AT$_n$-DE, respectively.

\section{Explanation Heatmaps}
\label{sec::Heatmaps}

\begin{figure}
	\centering
	\subfloat[AT$_c$-AT$_c$]{
		\includegraphics[width=0.235\linewidth]{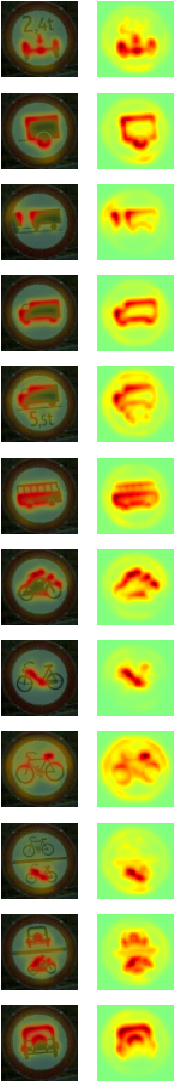}
	}
	\quad
	\subfloat[AT$_n$-AT$_n$]{
		\includegraphics[width=0.235\linewidth]{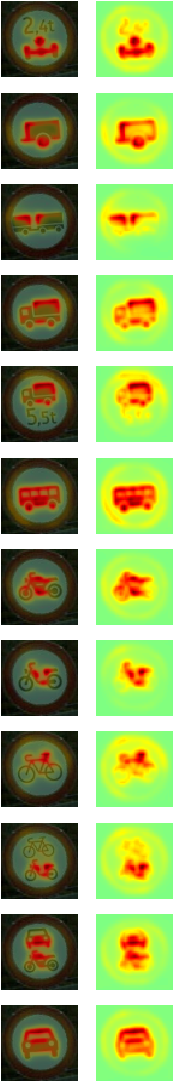}
	}
	\quad
	\subfloat[DE-DE]{
		\includegraphics[width=0.235\linewidth]{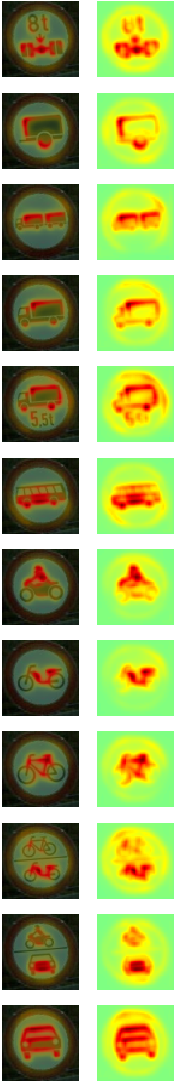}
	}
	\caption{Average explanations of all correctly predicted test images of classes 1--12.}
	\label{fig::correct_own_all_1}
\end{figure}

\begin{figure}
	\centering
	\subfloat[AT$_c$-AT$_c$]{
		\includegraphics[width=0.235\linewidth]{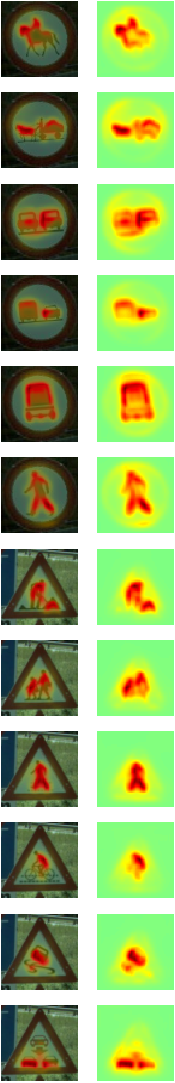}
	}
	\quad
	\subfloat[AT$_n$-AT$_n$]{
		\includegraphics[width=0.235\linewidth]{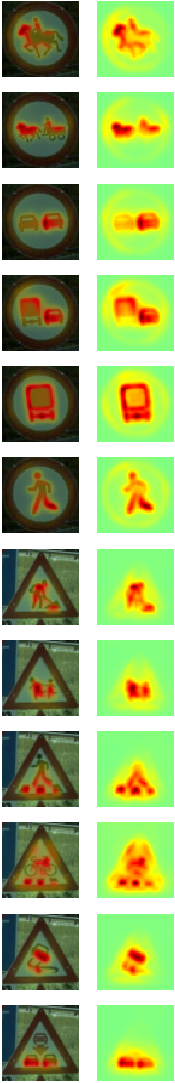}
	}
	\quad
	\subfloat[DE-DE]{
		\includegraphics[width=0.235\linewidth]{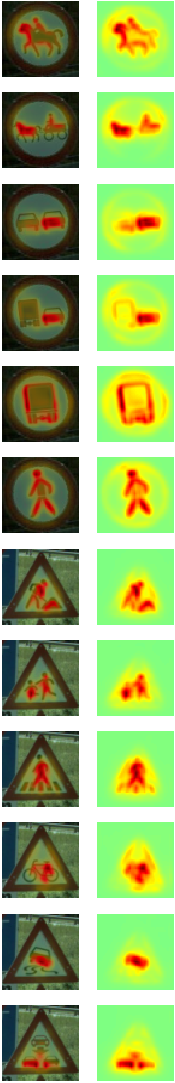}
	}
	\caption{Average explanations of all correctly predicted test images of classes 13--24.}
	\label{fig::correct_own_all_2}
\end{figure}

Figures~\ref{fig::correct_own_all_1} and~\ref{fig::correct_own_all_2} show the average LRP explanations of all correctly predicted test images, for the models trained on the whole data set.

\section{Results for ResNet}
\label{sec::ResNet}

Since the main part of the paper focuses on the results of the Li-Wang models, we present here detailed results for the ResNet models.

\begin{table}[t]
	\centering
	\begin{tabular}{l@{\qquad}rrrrrr}
		& 1 & 2 & 3 & 4 & 5 & All\\\hline\hline
		AT$_c$-AT$_c$ & \textbf{98.42} & 97.65 & 97.03 & \textbf{95.67} & 93.53 & \textbf{98.48}\\
		AT$_n$-AT$_n$ & 98.35 & 98.21 & 96.64 & 95.49 & 94.13 & 98.45\\
		DE-DE & 97.29 & \textbf{98.57} & \textbf{97.14} & 95.39 & \textbf{94.44} & 98.23\\\hline
		AT$_c$-AT$_n$ & 68.85 & 63.32 & 72.42 & 68.46 & 68.83 & 77.76\\
		AT$_c$-DE & 78.04 & 72.54 & 77.82 & 77.32 & 73.63 & 80.43\\
		AT$_n$-DE & 73.06 & 74.04 & 74.39 & 73.42 & 71.65 & 74.24\\
		DE-AT$_c$ & 67.78 & 70.67 & 70.22 & 73.17 & 71.64 & 77.26\\
		DE-AT$_n$ & 67.11 & 73.58 & 72.74 & 72.35 & 69.11 & 72.82
	\end{tabular}
	\caption{Classification accuracy (\%) of the ResNet models trained in our experiments. Columns correspond to corruption intensity levels. Shown is the average accuracy over three runs, with the top accuracy per corruption intensity in \textbf{bold}.}
	\label{tab::acc_resnet}
\end{table}

\begin{figure}
	\centering
	\footnotesize
	\begin{tabular}{cc}
		\includegraphics[width=0.45\linewidth]{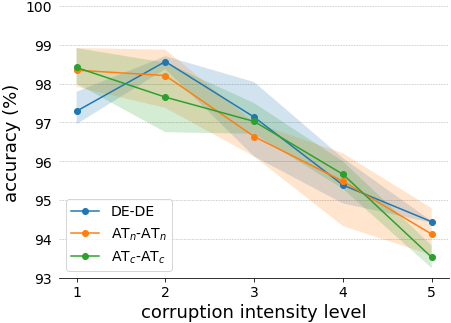} & \includegraphics[width=0.45\linewidth]{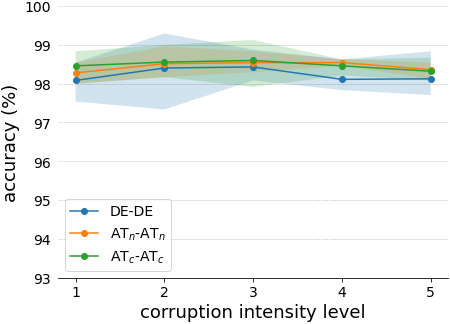}\\
		(a) `own' test sets, separate models per intensity & (b) `own' test sets, one model for all intensities\\[1em]
		\includegraphics[width=0.45\linewidth]{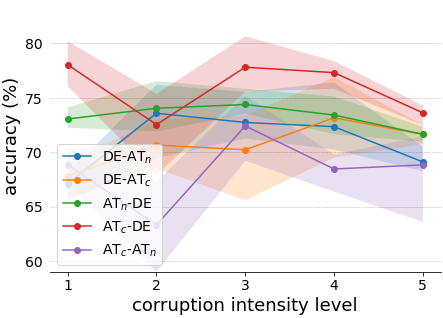} & \includegraphics[width=0.45\linewidth]{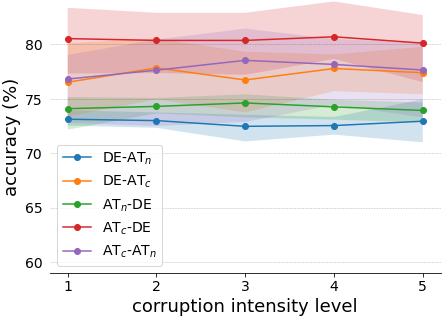}\\
		(c) `foreign' test sets, separate models per intensity & (d) `foreign' test sets, one model for all intensities
	\end{tabular}
	\caption{Classification accuracy plotted against corruption intensity, for the ResNet models. The semi-transparent areas indicate the minimum/maximum over the three runs. In (b) and (d), even though the models were trained on the full data set with all corruption intensities, the classification accuracy is again shown for each corruption intensity.}
	\label{fig::acc_resnet}
\end{figure}

Comparing Table~\ref{tab::acc_resnet} to Table~\ref{tab::acc} and Figure~\ref{fig::acc_resnet} to Figure~\ref{fig::acc}, one can see that ResNet performs slightly worse than Li-Wang in our experiments. The general tendency is the same as with Li-Wang, though: if models are evaluated on the pictogram design group they were trained on the design hardly matters (Figure~\ref{fig::acc_resnet} (b)), and if they are evaluated on different designs models trained on current Austrian pictograms generalize better to German pictograms than models trained on proposed new Austrian pictograms, and vice versa (Figure~\ref{fig::acc_resnet} (d)). The main difference between Li-Wang and ResNet is that in the latter case the German models perform slightly worse than the two Austrian models.

\begin{table}
	\centering
	\small
	\begin{tabular}{l|rrrrrrr}
		&  All &   Per-level rank &     1 &    2 &    3 &    4 &    5 \\\hline
		\Fahrrad              & 99.40 &   4.1 ($\pm$8.6) & 99.33 & 99.67 &  98.00 & 98.33 & 98.33 \\
		\Baustelle            & 99.33 &   9.5 ($\pm$7.0) & 99.00 & 96.00 &  98.33 & 97.67 & 95.67 \\
		\Achslast             & 99.33 &   4.4 ($\pm$8.0) & 99.00 & 99.33 &  99.00 & 99.00 & 97.33 \\
		\FahrradMoped         & 99.13 &  10.6 ($\pm$5.5) & 99.00 & 97.33 &  96.33 & 97.00 & 97.33 \\
		\GewichtLKW           & 99.13 &   4.3 ($\pm$8.1) & 99.33 & 99.67 & 100.00 & 97.33 & 95.33 \\
		\Fussgaenger          & 99.07 &   6.7 ($\pm$7.9) & 98.33 & 99.33 &  98.33 & 99.33 & 95.67 \\
		\Reitverbot           & 99.07 &  14.9 ($\pm$5.5) & 99.33 & 95.67 &  97.00 & 94.33 & 94.00 \\
		\Gefahrengut          & 99.07 &   4.7 ($\pm$7.1) & 99.67 & 99.00 &  99.00 & 96.67 & 96.00 \\
		\Anhaenger            & 99.00 &   9.6 ($\pm$6.0) & 99.67 & 96.67 &  97.67 & 97.00 & 94.33 \\
		\Kinder               & 98.93 &  15.6 ($\pm$5.5) & 96.33 & 97.00 &  97.33 & 94.33 & 94.67 \\
		\Fuhrwerk             & 98.53 &  15.2 ($\pm$6.2) & 98.33 & 98.00 &  93.00 & 95.00 & 95.33 \\
		\Fahrradueberfahrt    & 98.53 &   9.1 ($\pm$5.2) & 99.33 & 97.67 &  98.33 & 97.00 & 94.00 \\
		\Motorrad             & 98.47 &  10.6 ($\pm$5.7) & 99.33 & 98.67 &  98.67 & 94.33 & 92.67 \\
		\KFZ                  & 98.40 &  19.4 ($\pm$7.8) & 95.33 & 96.00 &  96.00 & 96.00 & 89.33 \\
		\Einspurige           & 98.20 &  10.1 ($\pm$3.8) & 98.67 & 99.00 &  98.00 & 96.00 & 94.67 \\
		\Schleudergefahr      & 98.20 &  16.1 ($\pm$5.6) & 97.00 & 98.33 &  96.67 & 95.33 & 90.67 \\
		\AnhaengerLKW         & 98.00 &  12.6 ($\pm$5.4) & 99.33 & 98.67 &  97.00 & 96.00 & 89.67 \\
		\LKW                  & 97.93 &  16.6 ($\pm$5.6) & 97.00 & 97.33 &  97.00 & 93.67 & 94.00 \\
		\Omnibus              & 97.80 &  17.2 ($\pm$9.4) & 99.67 & 96.33 &  94.67 & 93.00 & 90.00 \\
		\Falschfahrer         & 97.80 &  14.0 ($\pm$2.4) & 99.00 & 97.33 &  97.33 & 95.33 & 92.33 \\
		\Ueberholverbot       & 97.73 &  15.8 ($\pm$6.7) & 98.33 & 99.00 &  95.67 & 93.00 & 94.00 \\
		\UeberholverbotLKW    & 97.67 &  15.1 ($\pm$4.4) & 98.67 & 97.67 &  95.00 & 96.33 & 92.33 \\
		\Fussgaengeruebergang & 97.53 &  22.0 ($\pm$8.9) & 94.67 & 94.33 &  96.00 & 91.00 & 90.67 \\
		\Moped                & 97.20 &  21.8 ($\pm$8.4) & 98.33 & 95.67 &  94.33 & 93.00 & 86.33
	\end{tabular}
	\caption{AT$_c$-AT$_c$ per-class accuracy (\%) for ResNet models, sorted descending by the accuracy in the models trained on the whole data set (column `All'). Per-level rank is the average ($\pm$ standard deviation) rank of each class for the models trained separately on the five corruption intensities (columns 1--5).}
	\label{tab::old_classes_resnet}
\end{table}

\begin{table}
	\centering
	\small
	\begin{tabular}{l|rrrrrrr}
		&  All &   Per-level rank &     1 &    2 &    3 &    4 &    5 \\\hline
		\Achslast             & 99.20 &  13.3 ($\pm$4.5) &  98.00 &  98.00 & 95.67 & 96.33 & 95.33 \\
		\Gefahrengut          & 99.20 &  10.9 ($\pm$5.5) &  98.33 &  99.33 & 95.00 & 96.33 & 95.33 \\
		\Baustelle            & 99.13 &  10.4 ($\pm$5.8) &  98.67 &  97.33 & 96.67 & 98.00 & 95.00 \\
		\Fussgaengeruebergang & 99.00 &   6.7 ($\pm$7.5) &  99.33 &  98.00 & 98.67 & 97.33 & 96.33 \\
		\Schleudergefahr      & 98.93 &   3.7 ($\pm$8.9) &  98.67 & 100.00 & 98.67 & 98.00 & 97.33 \\
		\Falschfahrer         & 98.87 &  12.3 ($\pm$5.7) &  98.33 &  99.67 & 96.67 & 96.00 & 92.33 \\
		\UeberholverbotLKW    & 98.87 &   7.3 ($\pm$6.2) &  99.00 &  99.33 & 97.00 & 98.00 & 94.33 \\
		\Einspurige           & 98.87 &  10.2 ($\pm$4.5) &  99.33 &  99.00 & 97.00 & 96.00 & 93.67 \\
		\Kinder               & 98.80 &  15.3 ($\pm$7.1) &  97.00 &  99.00 & 93.33 & 96.33 & 94.33 \\
		\Fussgaenger          & 98.80 &  12.4 ($\pm$8.4) &  97.33 &  97.00 & 95.67 & 98.00 & 96.67 \\
		\Fuhrwerk             & 98.80 &  16.6 ($\pm$5.9) &  98.00 &  97.67 & 93.67 & 95.33 & 94.67 \\
		\LKW                  & 98.80 &  11.7 ($\pm$5.4) &  98.33 &  98.67 & 99.33 & 94.33 & 94.00 \\
		\FahrradMoped         & 98.73 &   5.0 ($\pm$7.8) &  99.67 &  99.00 & 99.33 & 96.67 & 96.00 \\
		\GewichtLKW           & 98.60 &   8.0 ($\pm$6.4) &  99.33 &  99.33 & 98.33 & 97.00 & 93.33 \\
		\Fahrradueberfahrt    & 98.60 &   9.1 ($\pm$7.4) & 100.00 &  98.33 & 99.33 & 94.67 & 94.33 \\
		\Fahrrad              & 98.47 &   5.5 ($\pm$7.3) &  99.67 &  99.67 & 98.33 & 97.00 & 94.67 \\
		\Omnibus              & 98.07 &  19.4 ($\pm$6.9) &  98.00 &  98.00 & 94.33 & 91.67 & 92.67 \\
		\Reitverbot           & 98.07 &  16.9 ($\pm$5.7) &  98.33 &  96.00 & 96.00 & 93.67 & 94.00 \\
		\Anhaenger            & 98.07 &  17.7 ($\pm$6.4) &  97.67 &  98.33 & 96.67 & 93.33 & 92.00 \\
		\AnhaengerLKW         & 98.07 &  15.4 ($\pm$5.0) &  98.33 &  96.33 & 97.00 & 95.33 & 93.00 \\
		\Ueberholverbot       & 97.93 &   9.3 ($\pm$4.5) &  99.00 &  98.67 & 97.00 & 95.67 & 96.00 \\
		\Moped                & 97.67 &  22.1 ($\pm$8.8) &  95.67 &  96.00 & 95.33 & 92.00 & 91.00 \\
		\KFZ                  & 97.33 &  21.0 ($\pm$8.2) &  97.33 &  95.67 & 95.67 & 93.67 & 90.33 \\
		\Motorrad             & 95.87 &  19.8 ($\pm$8.2) &  97.00 &  98.67 & 94.67 & 91.00 & 92.33
	\end{tabular}
	\caption{AT$_n$-AT$_n$ per-class accuracy (\%) for ResNet models, sorted descending by the accuracy in the models trained on the whole data set (column `All'). Per-level rank is the average ($\pm$ standard deviation) rank of each class for the models trained separately on the five corruption intensities (columns 1--5).}
	\label{tab::new_classes_resnet}
\end{table}

\begin{table}
	\centering
	\small
	\begin{tabular}{l|rrrrrrr}
		&  All &   Per-level rank &     1 &    2 &    3 &    4 &    5 \\\hline
		\Fussgaengeruebergang & 99.53 &   4.6 ($\pm$7.7) &  98.33 &  99.33 & 99.67 & 98.00 & 96.67 \\
		\Gefahrengut          & 99.27 &   4.0 ($\pm$8.8) &  99.67 & 100.00 & 97.33 & 98.00 & 98.33 \\
		\UeberholverbotLKW    & 99.20 &   8.0 ($\pm$5.6) &  99.00 &  99.33 & 97.33 & 96.33 & 96.33 \\
		\Einspurige           & 98.93 &   8.0 ($\pm$6.8) &  99.67 &  98.33 & 97.33 & 96.33 & 96.67 \\
		\GewichtLKW           & 98.87 &   8.9 ($\pm$6.2) &  98.67 &  99.67 & 98.67 & 95.00 & 94.00 \\
		\FahrradMoped         & 98.87 &   3.9 ($\pm$8.5) &  99.67 &  99.33 & 98.67 & 99.00 & 96.67 \\
		\Fahrrad              & 98.80 &  11.0 ($\pm$5.7) &  99.33 &  99.00 & 96.33 & 96.67 & 94.00 \\
		\Fuhrwerk             & 98.73 &  17.1 ($\pm$7.8) &  95.00 &  97.33 & 96.33 & 95.00 & 96.33 \\
		\Baustelle            & 98.73 &   8.4 ($\pm$4.4) &  98.00 &  99.33 & 97.67 & 96.33 & 95.67 \\
		\Fussgaenger          & 98.73 &  12.8 ($\pm$6.6) &  97.67 &  98.00 & 95.33 & 98.00 & 95.33 \\
		\Achslast             & 98.73 &   9.8 ($\pm$6.1) &  95.67 &  99.67 & 98.00 & 96.00 & 95.33 \\
		\Omnibus              & 98.60 &  15.5 ($\pm$7.4) &  97.33 &  99.67 & 95.33 & 93.33 & 92.67 \\
		\Schleudergefahr      & 98.33 &  13.6 ($\pm$7.6) &  98.00 &  97.00 & 99.00 & 92.67 & 94.33 \\
		\Fahrradueberfahrt    & 98.13 &  12.8 ($\pm$4.4) &  96.00 &  98.33 & 97.00 & 97.00 & 94.33 \\
		\Kinder               & 97.93 &  14.9 ($\pm$3.4) &  97.33 &  98.00 & 97.33 & 95.00 & 93.67 \\
		\Falschfahrer         & 97.87 &  14.6 ($\pm$7.5) &  98.00 &  99.67 & 96.67 & 93.67 & 91.33 \\
		\KFZ                  & 97.67 &  20.4 ($\pm$9.2) &  91.33 &  98.67 & 92.67 & 92.33 & 92.67 \\
		\LKW                  & 97.60 &  14.1 ($\pm$6.8) &  95.33 &  98.00 & 99.00 & 96.00 & 92.67 \\
		\Reitverbot           & 97.47 &  11.7 ($\pm$4.2) &  96.67 &  98.33 & 97.67 & 95.33 & 96.33 \\
		\Motorrad             & 97.47 &  17.2 ($\pm$7.6) &  96.67 &  99.33 & 95.33 & 92.33 & 93.33 \\
		\Moped                & 97.33 &  20.0 ($\pm$8.4) &  95.00 &  96.33 & 96.67 & 95.67 & 92.00 \\
		\AnhaengerLKW         & 97.27 &  18.3 ($\pm$7.1) &  95.33 &  98.00 & 97.33 & 93.00 & 92.33 \\
		\Anhaenger            & 97.20 &  14.2 ($\pm$4.9) &  97.33 &  97.33 & 97.67 & 96.00 & 93.33 \\
		\Ueberholverbot       & 96.27 &  16.2 ($\pm$9.0) & 100.00 &  97.67 & 97.00 & 92.33 & 92.33
	\end{tabular}
	\caption{DE-DE per-class accuracy (\%) for ResNet models, sorted descending by the accuracy in the models trained on the whole data set (column `All'). Per-level rank is the average ($\pm$ standard deviation) rank of each class for the models trained separately on the five corruption intensities (columns 1--5).}
	\label{tab::DE_classes_resnet}
\end{table}

Tables~\ref{tab::old_classes_resnet}, \ref{tab::new_classes_resnet} and~\ref{tab::DE_classes_resnet} list the per-class accuracies of the ResNet models.

\begin{figure}
	\centering
	\includegraphics[width=\linewidth]{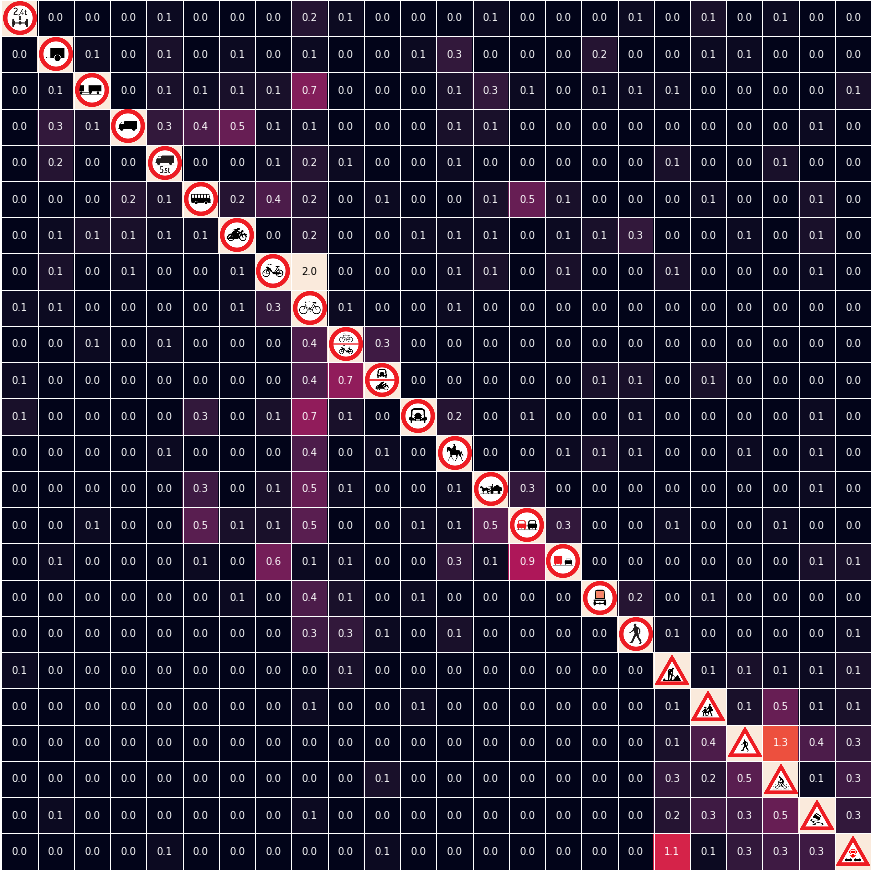}
	\caption{AT$_c$-AT$_c$ confusion matrix for the ResNet models trained on the whole data set. Rows represent true classes, columns represent predictions, numbers are percentages of misclassified samples (sum up to 100 row-wise).}
	\label{fig::confusion_old_all_resnet}
\end{figure}

\begin{figure}
	\centering
	\includegraphics[width=\linewidth]{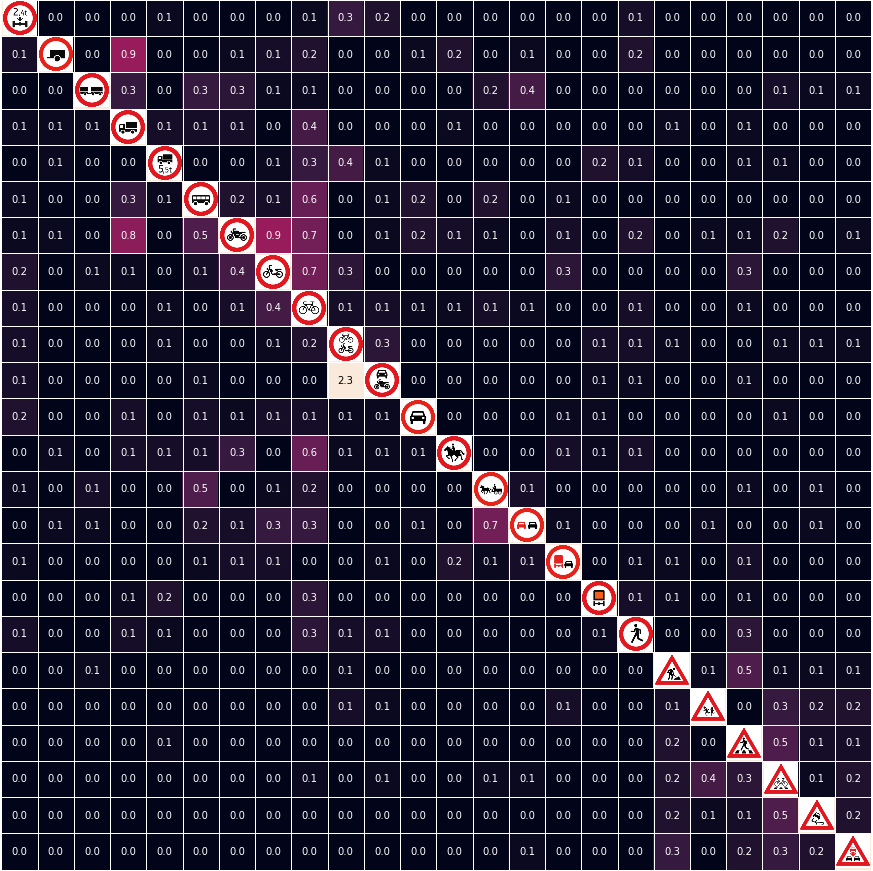}
	\caption{AT$_n$-AT$_n$ confusion matrix for the ResNet models trained on the whole data set. Rows represent true classes, columns represent predictions, numbers are percentages of misclassified samples (sum up to 100 row-wise).}
	\label{fig::confusion_new_all_resnet}
\end{figure}

\begin{figure}
	\centering
	\includegraphics[width=\linewidth]{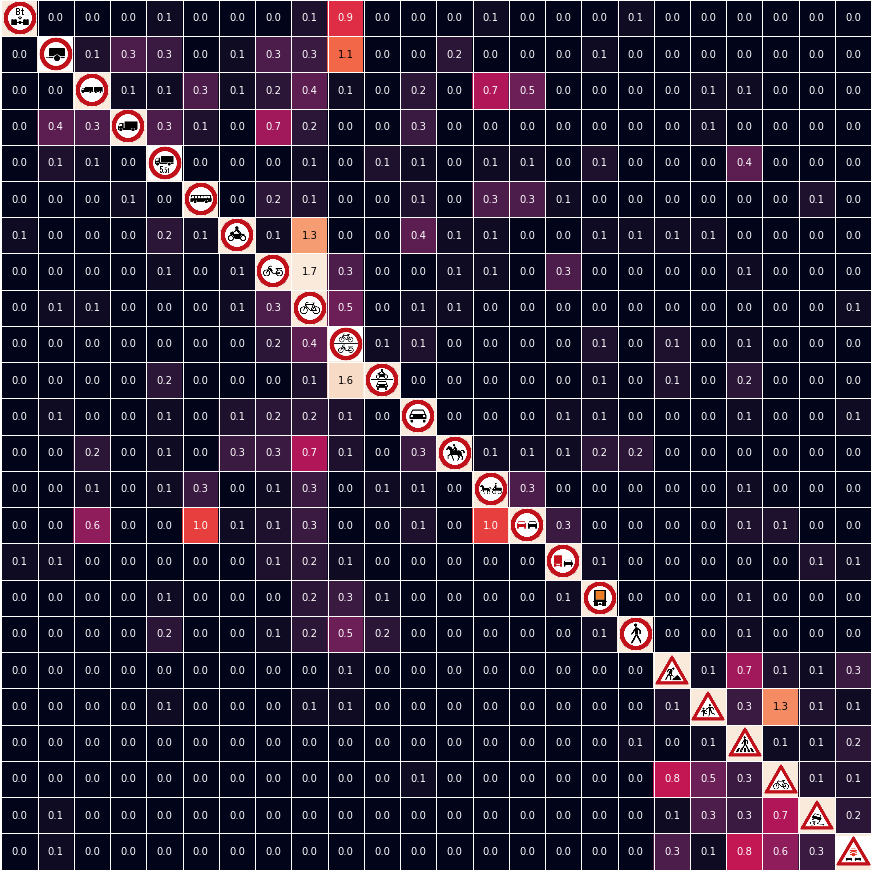}
	\caption{DE-DE confusion matrix for the ResNet models trained on the whole data set. Rows represent true classes, columns represent predictions, numbers are percentages of misclassified samples (sum up to 100 row-wise).}
	\label{fig::confusion_DE_all_resnet}
\end{figure}

Figures~\ref{fig::confusion_old_all_resnet}, \ref{fig::confusion_new_all_resnet} and~\ref{fig::confusion_DE_all_resnet} show the confusion matrices for the ResNet models for AT$_c$-AT$_c$, AT$_n$-AT$_n$ and DE-DE, respectively. Comparing them with the corresponding confusion matrices for the Li-Wang models (Figures~\ref{fig::confusion_old_all}, \ref{fig::confusion_new_all} and~\ref{fig::confusion_DE_all}) one can see that both model architectures frequently confuse the same classes.

\begin{figure}
	\centering
	\includegraphics[width=\linewidth]{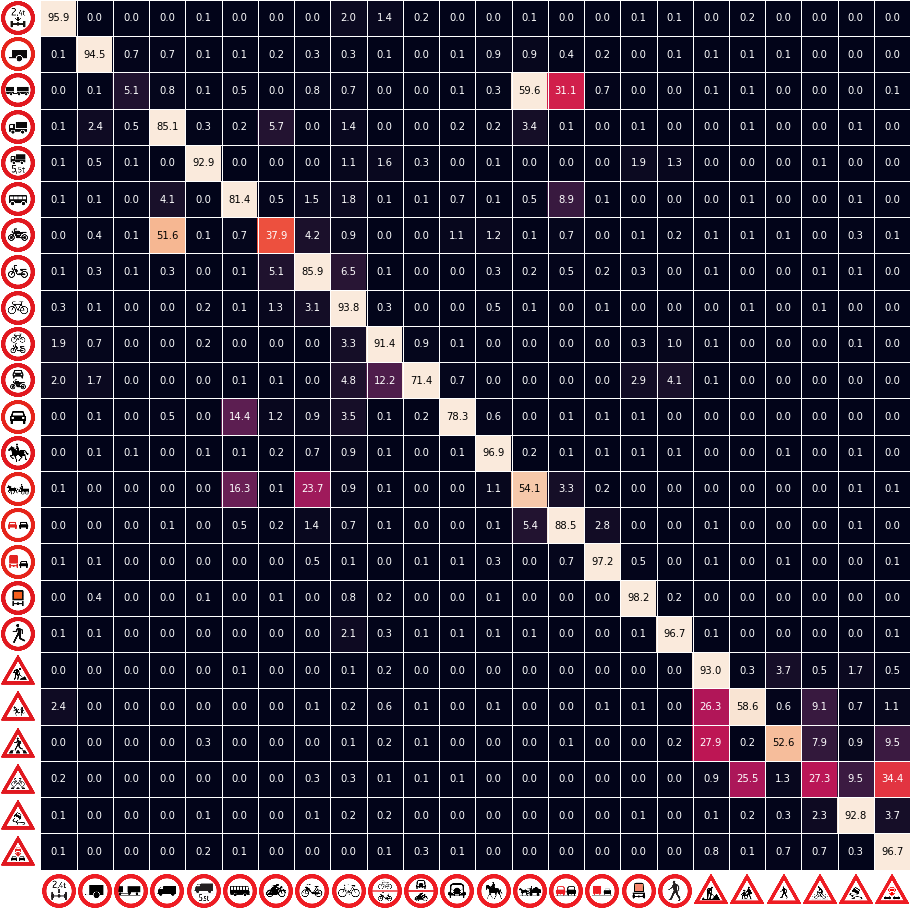}
	\caption{AT$_c$-AT$_n$ confusion matrix for the ResNet models trained on the whole data set. Rows represent true classes, columns represent predictions, numbers are percentages of misclassified samples (sum up to 100 row-wise).}
	\label{fig::confusion_old_new_all_resnet}
\end{figure}

\begin{figure}
	\centering
	\includegraphics[width=\linewidth]{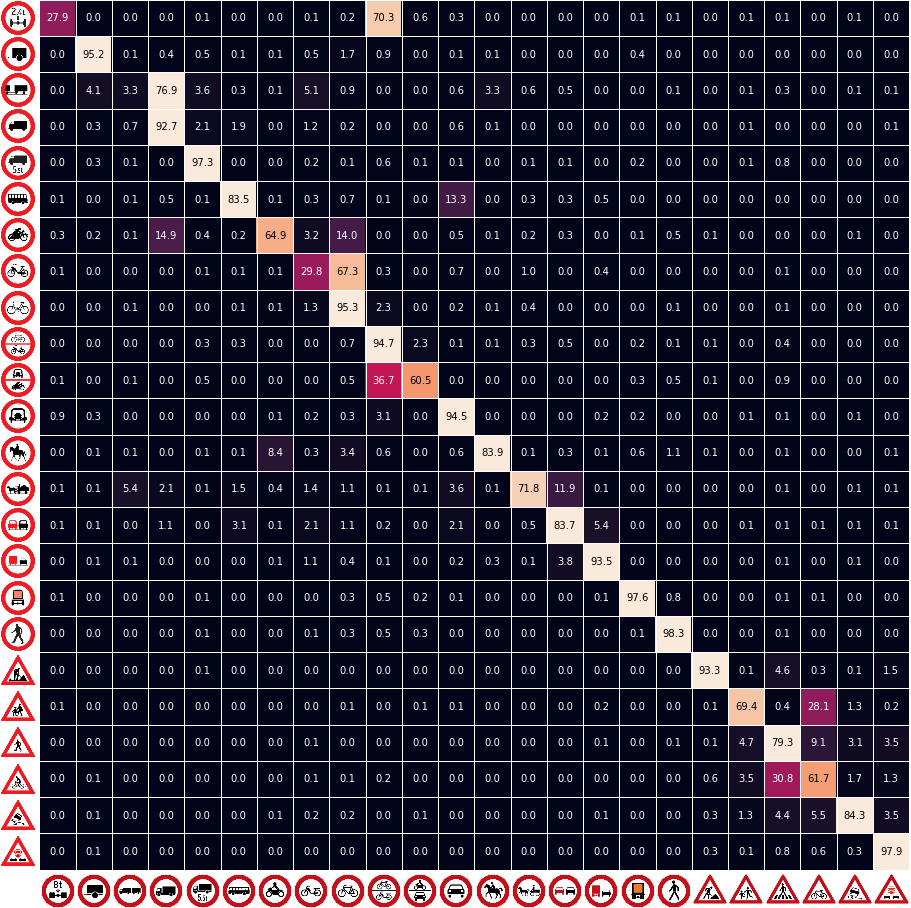}
	\caption{DE-AT$_c$ confusion matrix for the ResNet models trained on the whole data set. Rows represent true classes, columns represent predictions, numbers are percentages of misclassified samples (sum up to 100 row-wise).}
	\label{fig::confusion_DE_old_all_resnet}
\end{figure}

\begin{figure}
	\centering
	\includegraphics[width=\linewidth]{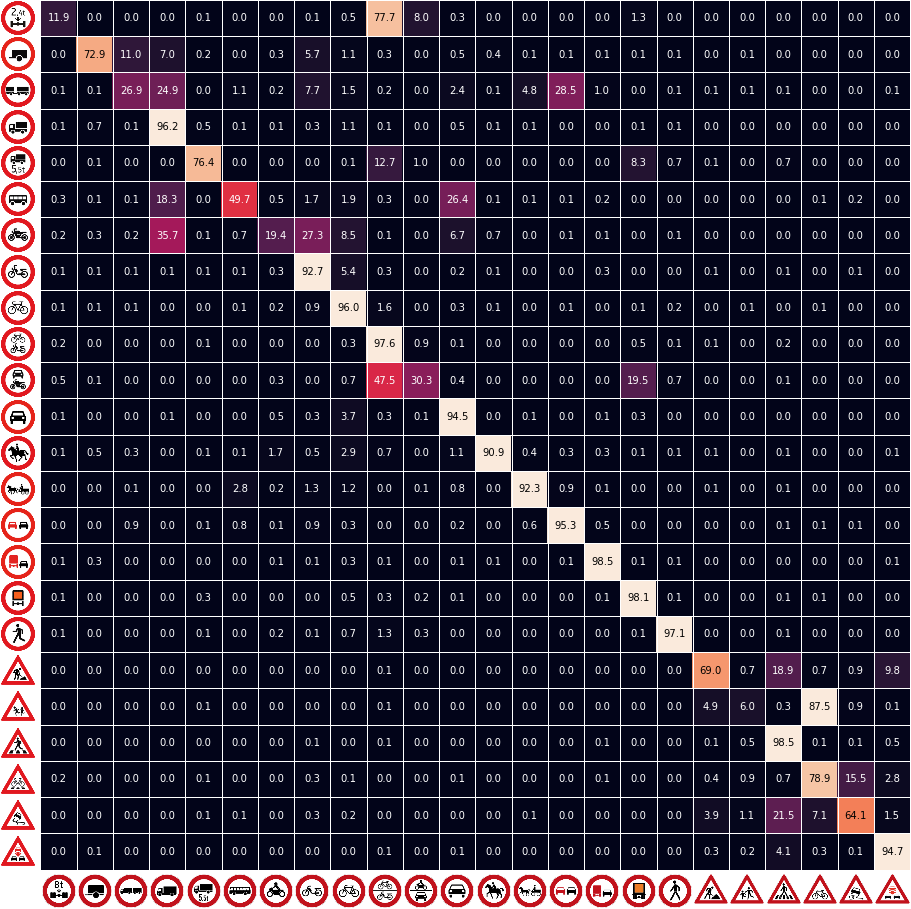}
	\caption{DE-AT$_n$ confusion matrix for the ResNet models trained on the whole data set. Rows represent true classes, columns represent predictions, numbers are percentages of misclassified samples (sum up to 100 row-wise).}
	\label{fig::confusion_DE_new_all_resnet}
\end{figure}

\begin{figure}
	\centering
	\includegraphics[width=\linewidth]{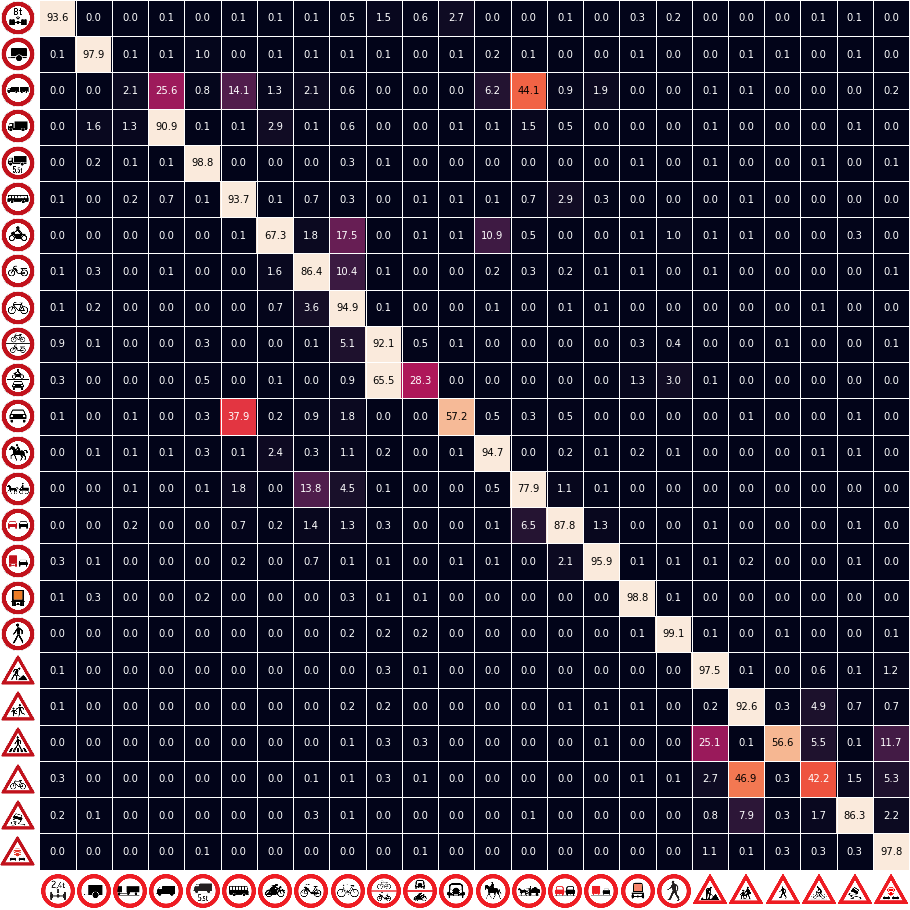}
	\caption{AT$_c$-DE confusion matrix for the ResNet models trained on the whole data set. Rows represent true classes, columns represent predictions, numbers are percentages of misclassified samples (sum up to 100 row-wise).}
	\label{fig::confusion_old_DE_all_resnet}
\end{figure}

\begin{figure}
	\centering
	\includegraphics[width=\linewidth]{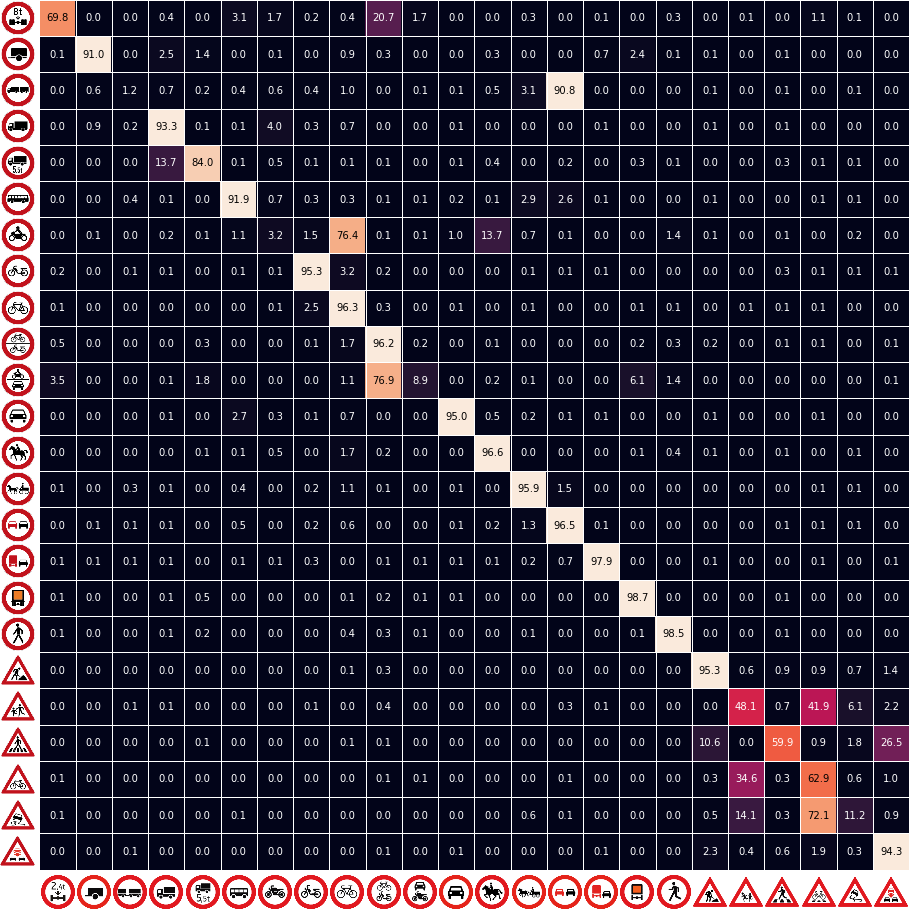}
	\caption{AT$_n$-DE confusion matrix for the ResNet models trained on the whole data set. Rows represent true classes, columns represent predictions, numbers are percentages of misclassified samples (sum up to 100 row-wise).}
	\label{fig::confusion_new_DE_all_resnet}
\end{figure}

Figures~\ref{fig::confusion_old_new_all_resnet}, \ref{fig::confusion_DE_old_all_resnet}, \ref{fig::confusion_DE_new_all_resnet}, \ref{fig::confusion_old_DE_all_resnet} and~\ref{fig::confusion_new_DE_all_resnet} show the confusion matrices for the ResNet models for AT$_c$-AT$_n$, DE-AT$_c$, DE-AT$_n$, AT$_c$-DE and AT$_n$-DE, respectively. Comparing them with the corresponding confusion matrices for the Li-Wang models (Figures~\ref{fig::confusion_old_new_all}, \ref{fig::confusion_DE_old_all}, \ref{fig::confusion_DE_new_all}, \ref{fig::confusion_old_DE_all} and~\ref{fig::confusion_new_DE_all}) one can see that both model architectures frequently confuse the same classes.

\section{Results for Models Trained on Multiple Designs}
\label{sec::Mixed}

\begin{table}[t]
	\centering
	\begin{tabular}{l@{\qquad}l@{\qquad}rrrr}
		& & Training designs & AT$_c$ & AT$_n$ & DE\\\hline\hline
		Li-Wang & CUR & 98.69 & 98.67 & 85.48 & 98.71\\
		& ALL & 98.43 & 98.48 & 98.30 & 98.51\\\hline
		ResNet & CUR & 98.28 & 98.27 & 84.17 & 98.30\\
		& ALL & 97.40 & 97.51 & 97.33 & 97.36
	\end{tabular}
	\caption{Classification accuracy (\%) of the models trained on several pictogram designs. `Training designs' is the accuracy on the design groups the models were trained on, i.\,e., in the case of CUR it is the average of columns `AT$_c$' and `DE', and in the case of ALL it is the average of the other three columns.}
	\label{tab::acc_mixed}
\end{table}

In addition to the models presented so far, which were all trained on one single pictogram design, we also investigated how well models would perform when trained on multiple designs simultaneously. To that end we combined all currently available pictograms (AT$_c$, DE) into the larger set CUR, and combined \emph{all} pictogram designs into the set ALL. In either case we considered the complete data sets with all corruption intensities. The final test set accuracies are summarized in Table~\ref{tab::acc_mixed}. Again, the ResNet models perform slightly worse than the Li-Wang models, so in the remainder we focus on the latter.

It can be seen that the models trained on CUR generalize better to AT$_n$ (which is not part of CUR) than the models trained on its constituent designs individually; the accuracies are $85.48\%$ vs. $80.18\%$ and $77.35\%$, respectively. So, even though there is \emph{some} improvement in terms of generalizability, the accuracy of CUR-AT$_n$ is still significantly lower than the accuracy on CUR's own test set ($98.69\%$). Training on a more diverse set of pictogram designs thus helps, but is far from optimal. As expected, the models trained on all three pictogram designs perform best overall, constantly reaching an accuracy of well over $98\%$ on each of the three designs. However, when comparing only the accuracy on the designs the models were trained on, CUR performs slightly better at $98.69\%$ vs. $98.43\%$.

\begin{figure}
	\centering
	\includegraphics[width=\linewidth]{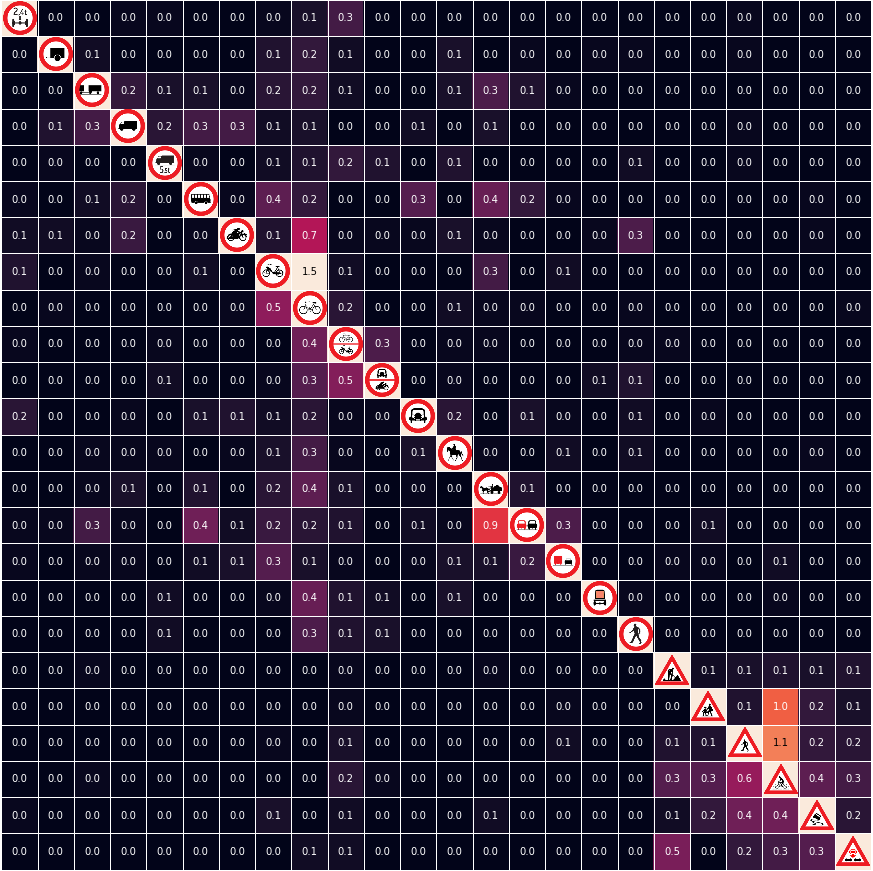}
	\caption{CUR-CUR confusion matrix for the Li-Wang models trained and evaluated on all corruption intensities. Rows represent true classes, columns represent predictions, numbers are percentages of misclassified samples (sum up to 100 row-wise).}
	\label{fig::confusion_CUR_CUR_all}
\end{figure}

\begin{figure}
	\centering
	\includegraphics[width=\linewidth]{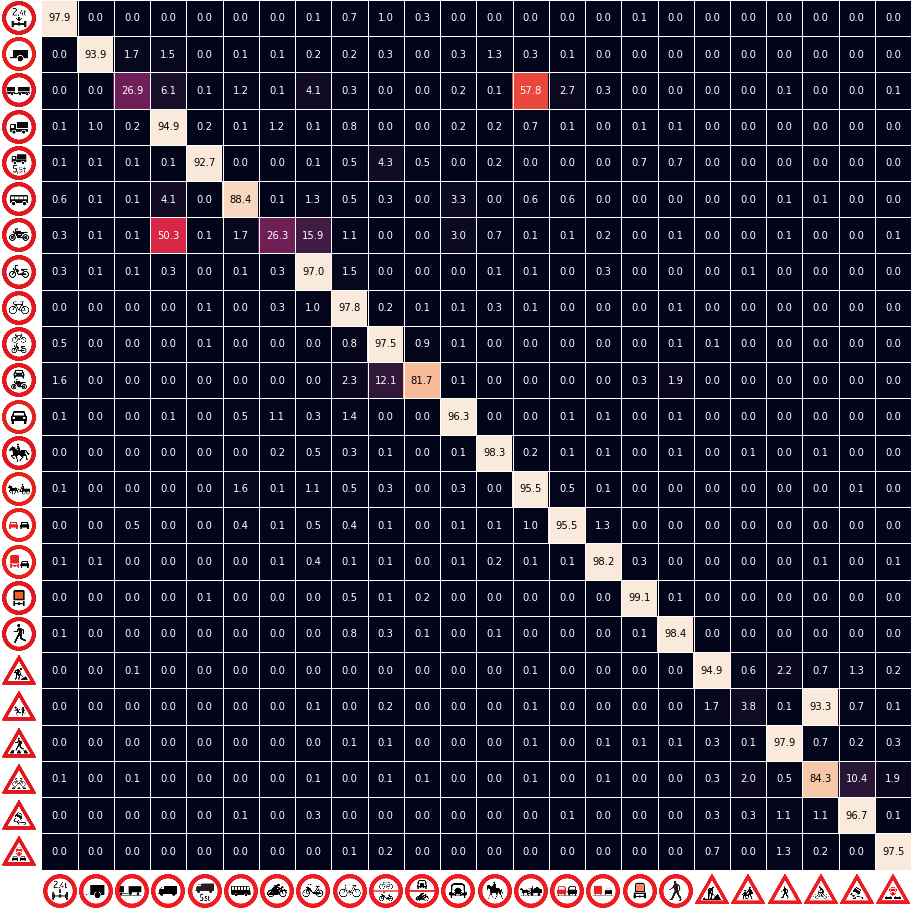}
	\caption{CUR-AT$_n$ confusion matrix for the Li-Wang models trained and evaluated on all corruption intensities. Rows represent true classes, columns represent predictions, numbers are percentages of misclassified samples (sum up to 100 row-wise).}
	\label{fig::confusion_CUR_new_all}
\end{figure}

\begin{figure}
	\centering
	\includegraphics[width=\linewidth]{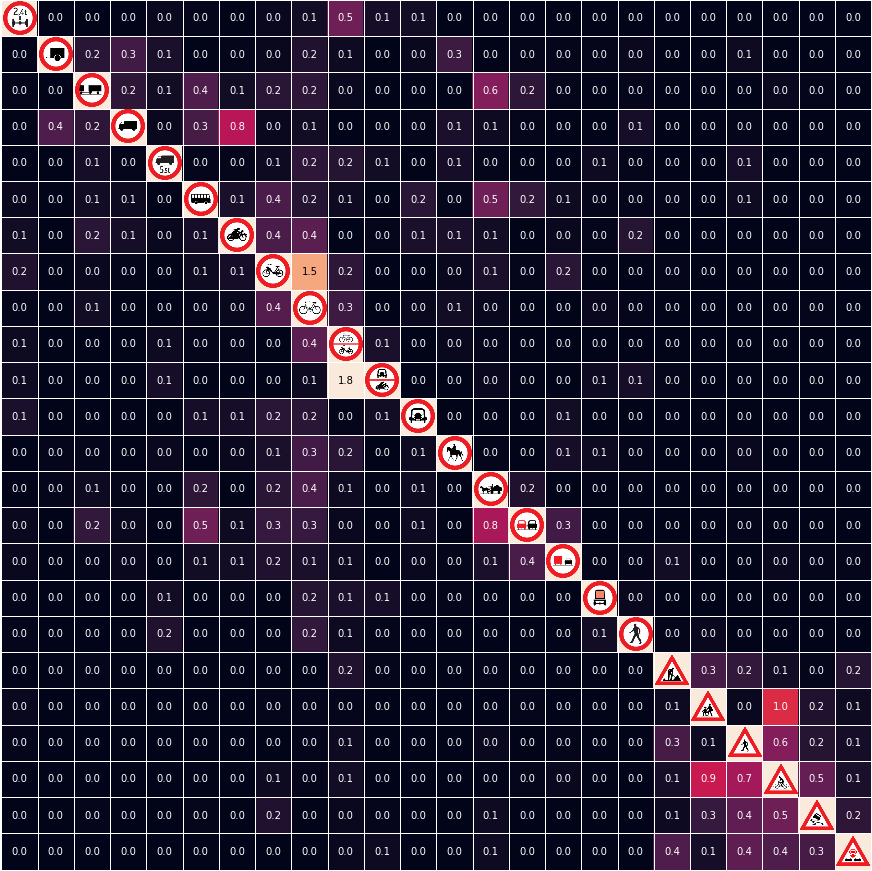}
	\caption{ALL-ALL confusion matrix for the Li-Wang models trained and evaluated on all corruption intensities. Rows represent true classes, columns represent predictions, numbers are percentages of misclassified samples (sum up to 100 row-wise).}
	\label{fig::confusion_ALL_ALL_all}
\end{figure}

Figures~\ref{fig::confusion_CUR_CUR_all}, \ref{fig::confusion_CUR_new_all} and~\ref{fig::confusion_ALL_ALL_all} show the full confusion matrices of the Li-Wang models for CUR-CUR, CUR-AT$_n$ and ALL-ALL, respectively. As can be seen there, the most frequently confused classes in CUR-AT$_n$ are:
\begin{itemize}
	\item `\Kinder' $\rightarrow$ `\Fahrradueberfahrt' ($93.3\%$)
	\item `\AnhaengerLKW' $\rightarrow$ `\Fuhrwerk' ($57.8\%$)
	\item `\Motorrad' $\rightarrow$ `\LKW' ($50.3\%$)
	\item `\Motorrad' $\rightarrow$ `\Moped' ($15.9\%$)
	\item `\KFZ' $\rightarrow$ `\FahrradMoped' ($12.1\%$)
	\item `\Fahrradueberfahrt' $\rightarrow$ `\Schleudergefahr' ($10.4\%$)
\end{itemize}
These results are more or less in line with those of AT$_c$-AT$_n$ and DE-AT$_n$ (cf. Table~\ref{tab::confusion_foreign}).

\begin{figure}
	\centering
	\subfloat[CUR-AT$_c$]{
		\includegraphics[width=0.235\linewidth]{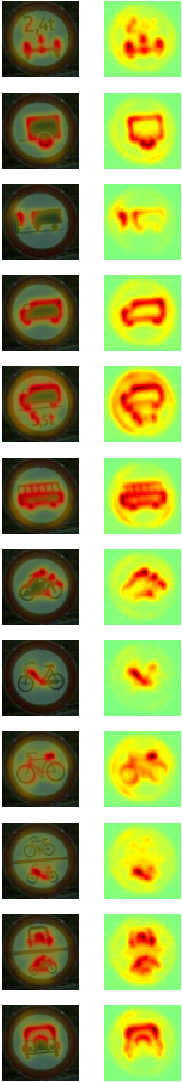}
	}
	\quad
	\subfloat[CUR-DE]{
		\includegraphics[width=0.235\linewidth]{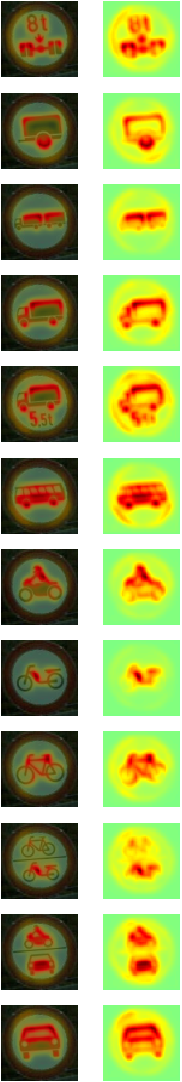}
	}
	\caption{Average explanations of all correctly predicted test images of classes 1--12.}
	\label{fig::correct_CUR_all_1}
\end{figure}

\begin{figure}
	\centering
	\subfloat[CUR-AT$_c$]{
		\includegraphics[width=0.235\linewidth]{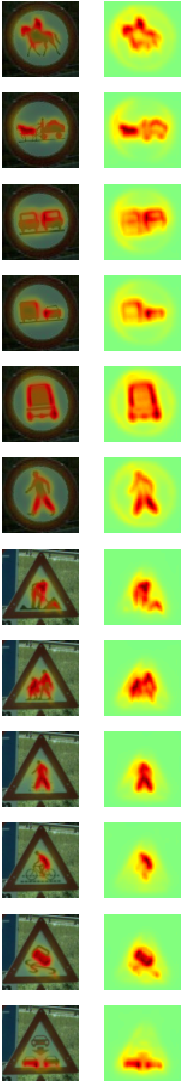}
	}
	\quad
	\subfloat[CUR-DE]{
		\includegraphics[width=0.235\linewidth]{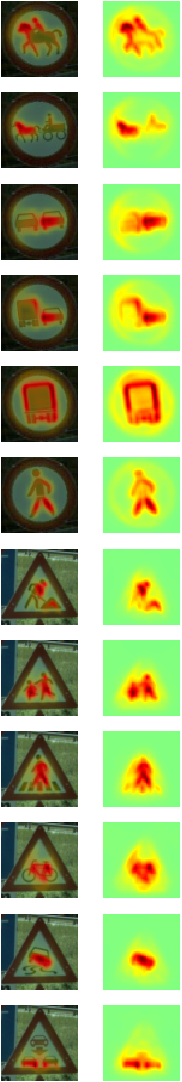}
	}
	\caption{Average explanations of all correctly predicted test images of classes 13--24.}
	\label{fig::correct_CUR_all_2}
\end{figure}

\begin{figure}
	\centering
	\subfloat[ALL-AT$_c$]{
		\includegraphics[width=0.235\linewidth]{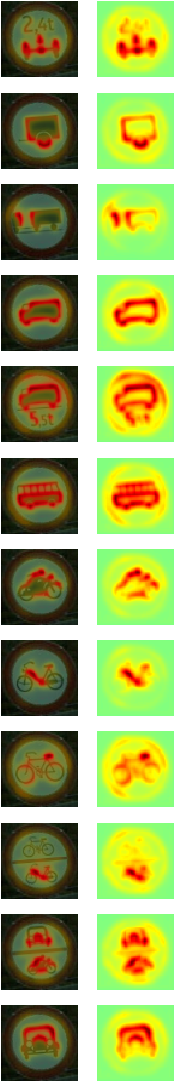}
	}
	\quad
	\subfloat[ALL-AT$_n$]{
		\includegraphics[width=0.235\linewidth]{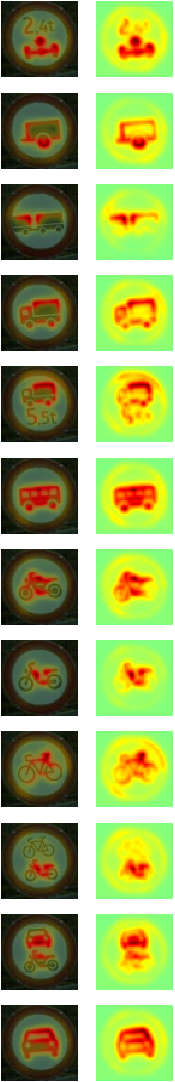}
	}
	\quad
	\subfloat[ALL-DE]{
		\includegraphics[width=0.235\linewidth]{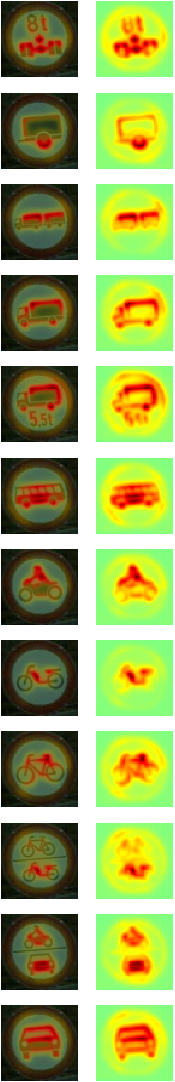}
	}
	\caption{Average explanations of all correctly predicted test images of classes 1--12.}
	\label{fig::correct_ALL_all_1}
\end{figure}

\begin{figure}
	\centering
	\subfloat[ALL-AT$_c$]{
		\includegraphics[width=0.235\linewidth]{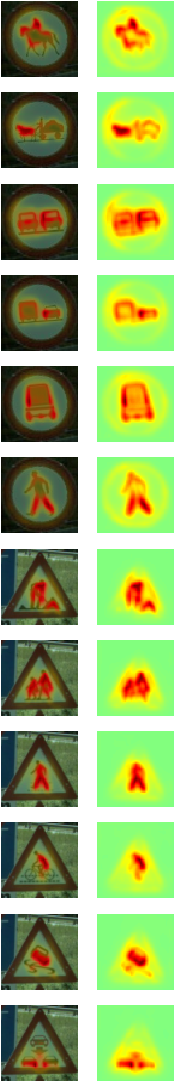}
	}
	\quad
	\subfloat[ALL-AT$_n$]{
		\includegraphics[width=0.235\linewidth]{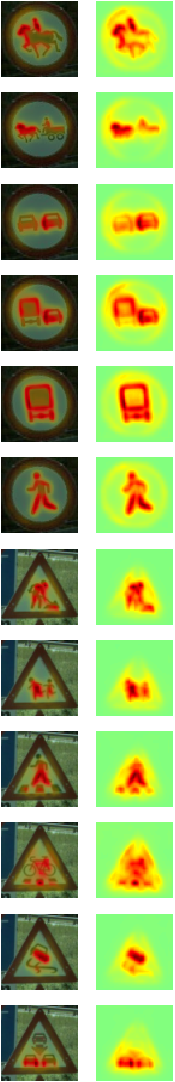}
	}
	\quad
	\subfloat[ALL-DE]{
		\includegraphics[width=0.235\linewidth]{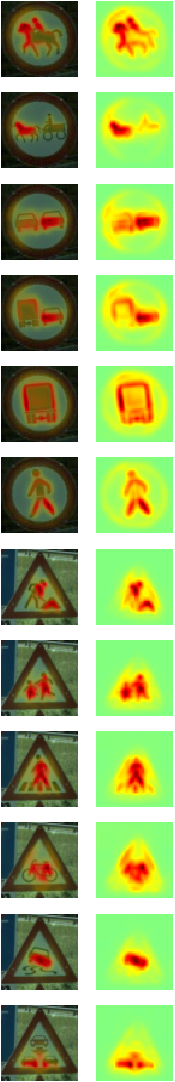}
	}
	\caption{Average explanations of all correctly predicted test images of classes 13--24.}
	\label{fig::correct_ALL_all_2}
\end{figure}

Figures~\ref{fig::correct_CUR_all_1}, \ref{fig::correct_CUR_all_2}, \ref{fig::correct_ALL_all_1} and~\ref{fig::correct_ALL_all_2} show the explanation heatmaps obtained from the CUR and ALL models, respectively. Comparing them to the corresponding heatmaps for the models trained on each pictogram design individually one can spot hardly any differences. This suggests that the models trained on multiple designs pay attention to the same image details, but learn that different patterns can be indicative of the same class.

\end{document}